\documentclass{article} % For LaTeX2e
\usepackage{iclr2020_conference,times}
\iclrfinalcopy

\usepackage{amsmath,amsfonts,bm}

\def\eqref#1{equation~\ref{#1}}
\def\1{\bm{1}}

\DeclareMathAlphabet{\mathsfit}{\encodingdefault}{\sfdefault}{m}{sl}
\SetMathAlphabet{\mathsfit}{bold}{\encodingdefault}{\sfdefault}{bx}{n}

\usepackage[hidelinks]{hyperref}
\usepackage{url}
\usepackage{graphicx}
\usepackage{nicefrac}
\usepackage{caption}
\usepackage{subcaption}

\newcommand{\code}{\texttt}

\title{Analysis of Video Feature Learning in Two-Stream CNNs on the Example of Zebrafish Swim Bout Classification}

\author{Bennet Breier \& Arno Onken \\
School of Informatics\\
The University of Edinburgh\\
Edinburgh, EH8 9AB, UK \\
\texttt{b.breier@sms.ed.ac.uk}, \texttt{aonken@inf.ed.ac.uk} \\
}

\begin{document}

\maketitle

\begin{abstract}
	% On its own, the first sentence is not clear - superior compared to what?.  You should start with a sentence on the Semmelhack results. Then you can say that deep neural network often reach superior performance when compared to SVMs. - Arno
	\cite{Semmelhack2014} have achieved high classification accuracy in distinguishing swim bouts of zebrafish using a Support Vector Machine (SVM).
	Convolutional Neural Networks (CNNs) have reached superior performance in various image recognition tasks over SVMs, but their learnt features are not immediately visible.
	Reaching better transparency helps to build trust in their classifications and makes learned features interpretable to experts.
	Using a recently developed technique called Deep Taylor Decomposition, we generated heatmaps to highlight input regions of high relevance for predictions.
	% In the previous sentence, you should emphasize that we beat the Semmelhack accuracy. - Arno
	We find that our CNN makes predictions by analyzing the steadiness of the tail's trunk, which markedly differs from the manually extracted features used by \cite{Semmelhack2014}.
	We further uncovered that the network paid attention to experimental artifacts.
	Removing these artifacts ensured the validity of predictions.
	After correction, our best CNN beats the SVM by 6.12\%, achieving a classification accuracy of 96.32\%.
	% The reference to the code should go into the Methods section. - Arno
	Our work thus demonstrates the utility of AI explainability for CNNs.
\end{abstract}

\section{Introduction} \label{sec:introduction}

% I think it would be good to start with some motivation based on the Semmelhack study.  You could highlight the importance of classification for tasks such as the one in Semmelhack, then explain that commonly SVMs were used for such tasks and then continue with the next sentence. - Arno
In the study by \cite{Semmelhack2014}, a well-performing classifier allowed to correlate neural interventions with behavioral changes.
Support Vector Machines (SVMs) were commonly applied to such classification tasks, relying on feature engineering by domain experts.
In recent years, Convolutional Neural Networks (CNNs) have proven to reach high accuracies in classification tasks on images and videos reducing the need for manual feature engineering.
After \cite{Lecun1995} introduced them in the 90s, CNNs had their break-through in the competition ILSVRC2012 with the architecture of \cite{Krizhevsky2012}.
Since then, more and more sophisticated architectures have been designed enabling them to identify increasingly abstract features.
This development has become possible due to the availability of larger training sets, computing resources, GPU training implementations, and better regularization techniques, such as Dropout (\cite{Hinton2012, Zeiler2014}).

While these more complex deep neural network architectures achieved better results, they also kept their learnt features hidden if not further analyzed.
This caused CNNs to come with significant drawbacks: a lack of trust in their classifications, missing interpretability of learned features in the application domain, and the absence of hints as to what data could enhance performance (\cite{Molnar2019}).
Explaining the decisions made by CNNs might even become a legal requirement in certain applications (\cite{Alber2018}).

In order to overcome these drawbacks, subsequent research has developed approaches to shed light on the inner workings of CNNs.
These approaches have been successfully used for uncovering how CNNs might learn unintended spurious correlations, termed ``Clever Hans'' predictions (\cite{Lapuschkin2019}).
Such predictions could even become harmful if the predictions entailed decisions with severe consequences (\cite{Leslie2019}).
Also, since deep neural networks have become a popular machine learning technique in applied domains, spurious correlations would undermine scientific discoveries.

This paper focuses on zebrafish research as an applied domain of AI explainability, considering that the research community around this organism has grown immensely.
The zebrafish is an excellent model organism for vertebrates, including humans, due to the following four reasons:
The genetic codes of humans and zebrafish are about 70\% orthologue (\cite{Howe2013}). 
The fish are translucent which allows non-invasive observation of changes in the organism (\cite{Bianco2011}). 
Furthermore, zebrafish are relatively cheap to maintain, produce plenty of offspring, and develop rapidly.
Finally, they are capable of recovering their brain structures within days after brain injury (\cite{Kishimoto2011, Kizil2012}).

In this paper, we adapt CNNs to work on highly controlled zebrafish video recordings and show the utility of a recently developed AI explainability technique on this task.
We train the network on optical flow for binary classifying swim bouts and achieve superior performance when compared to the current state-of-the-art in bout classification~(\cite{Semmelhack2014}).
We then create heatmaps over the videos with the ``iNNvestigate'' toolbox (\cite{Alber2018}) which highlight the areas that our CNN pays attention to when making a prediction.
The resulting heatmaps show that our CNN learns reasonable features which are very different from those manually composed by \cite{Semmelhack2014}.

\section{Related Work} \label{sec:related}

% A short introductory paragraph would be good here to improve the text flow. The paragraph should give an overview of the subsection in the section. - Arno

In the following, we will give an overview over relevant CNN architectures and approaches.
Then, we will summarize existing AI explainability approaches focusing on attribution techniques.
Finally, we highlight important studies of behavioral zebrafish research and give details of the study by \cite{Semmelhack2014}.

\textbf{CNN architectures.}
\cite{Carreira2017} identified five relevant types of video architectures:
CNN + LSTM (Long-Short-Term Memory), 3D-CNN (\cite{Ji2013}), Two-Stream (\cite{Simonyan2014}), 3D-Fused Two-Stream, and Two-Stream 3D-CNN.
They differ in whether the convolutions are based on 2D or 3D kernels, whether optical flow is added, and how consecutive frames exchange information.
Optical flow can be described as the horizontal and vertical displacement of a pixel from one frame to the next (\cite{Farneback2003}).
Several algorithms for flow calculation exist, such as TV-L1 (\cite{Zach2007}), Brox (\cite{Brox2004}), and Farneback (\cite{Farneback2003}).
Even novel deep learning approaches have been developed, e.g. FlowNet (\cite{Dosovitskiy2015, Ilg2017}).

\cite{Carreira2017} initialized several CNN architectures with pre-trained weights and trained them on human action recognition datasets to show the benefits of transfer learning in video classifiers.
Previous studies had shown this for images (\cite{Shin2016}).
Also, they found that adding a temporal stream based on optical flow always improved performance.
This idea of a network with a spatial stream for appearance and a temporal stream for motion had been first developed by \cite{Simonyan2014}.

\textbf{AI explainability techniques.}
Current AI explainability techniques on images can be largely categorized into two types:
attribution and feature visualization (\cite{Olah2017}).
Attribution relates regions of a sample image to activations of units in the CNN, while feature visualization uncovers what kinds of inputs strongly activate a particular unit.

One of the earlier and quite successful attribution approaches is called Sensitivity Analysis (\cite{Simonyan2014b}).
It is based on the idea that if a single pixel were marginally changed, the prediction would worsen significantly for an important pixel, but only slightly for less important ones.
Furthermore, \cite{Zeiler2014} and \cite{Ribeiro2016} showed simple ways of producing approximate relevance heatmaps.
By occluding parts of an image one by one and observing the change in activation, they could measure the influence of different image regions.
This allowed them to check whether the CNN focused on the important objects of the image or performed its classification only based on contextual information.

The technique applied in this paper is called Deep Taylor Decomposition (\cite{Montavon2017}), which arose from Layer-Wise Relevance Propagation (\cite{Bach2015}).
It has been put into use with text (\cite{Arras2017}), speech (\cite{Becker2018}), and only once with video data (\cite{Anders2018}).
It highlights the areas in a sample image which the CNN deems most relevant for making a correct classification.
We assume that the relevance of a pixel is determined by how much the classification accuracy would deteriorate if we altered this pixel. 
DTD distributes relevance from the output layer to the input layer by applying specific propagation rules for each layer.
This approach equals a first-order Taylor decomposition of the output function.
The authors argue that it yields a better approximation of relevance than Sensitivity Analysis.
\cite{Alber2018} included this technique in their ``iNNvestigate'' toolbox, which we used in our work.

Apart from attribution techniques, CNNs can be explained using direct feature visualization approaches.
They try to find intuitive representations of the patterns a given unit responds to particularly strongly or weakly.
Among these are deconvolution networks (\cite{Zeiler2014}), which are closely related to Sensitivity Analysis, and optimization approaches in the input domain (\cite{Erhan2009, Olah2018}).
Also, instead of creating an abstract representation of features, one can select specific samples which highly activate or suppress a particular unit.
\cite{Bau2017} even went a step further by hiding irrelevant parts of the input image, similar to LIME (\cite{Ribeiro2016}).

\textbf{Behavioral zebrafish research.}
As we argue in Section~\ref{sec:introduction}, zebrafish are a highly suitable model organism for the human body.
They serve as the object of study in many fields, such as wound repair (\cite{Kishimoto2011, Kizil2012}), visual processing in the brain (\cite{Roeser2003, Gahtan2005, Semmelhack2014, Temizer2015}), cancer research (\cite{White2013}), and genetic modifications (\cite{Hwang2013}).
Especially in neuroscientific research, understanding behavior and behavioral changes in response to cerebral interventions is of high importance (\cite{Krakauer2017}).

Previous studies therefore closely investigated the motion patterns of zebrafish (\cite{Borla2002, McElligott2005}).
\cite{Borla2002} described characteristic motion patterns during prey bouts, such as the precise bending of the very tip of the tail to bring the head into position and the subsequent strong arching of the tail's center.
Based on their observations, they concluded that zebrafish must possess fine axial motor control in all parts of their tail.

Given that prey movements can be clearly distinguished from other types of movements, \cite{Semmelhack2014} hypothesized that there must be a dedicated circuitry in the zebrafish brain.
They found a pathway from retinal ganglion cells to an area called AF7 projecting to the optic tectum, the nucleus of the medial longitudinal fasciculus, and the hindbrain, which in turn produces the characteristic motor output.
They verified their findings by ablating the AF7 neuropil and observing that lesioned fish failed to respond to prey stimuli with a movement that a trained SVM would classify as prey.
% The following should go into related work. - Arno
They have identified the following five features as the most discriminating ones, ordered by descending importance:
\begin{enumerate}
	\item Maximum tail curvature (maximum over the bout)
	\item Number of peaks in tail angle
	\item Mean tip angle (absolute value of tip angle in each frame, average over the bout)
	\item Maximum tail angle (maximum over the bout)
	\item Mean tip position (average position of last eight points in tail, with horizontal deflection as a fraction of the tail length)
\end{enumerate}
In our work, we make use of the dataset they gathered for training their SVM and compare our CNN to the features above.

\section{Methods} \label{sec:methods}

% Here, you should insert a short paragraph, first saying that we adapted CNNs to the task (describing the task again in half a sentence) and then giving a very short overview of the different steps. - Arno
We trained a Two-Stream CNN to distinguish prey and spontaneous swim bouts of larval zebrafish with short video samples.
For data preparation, we first extracted standardized snippets from raw videos and then performed augmentation by subsampling, flipping, and cropping.
After training, we computed heatmaps showing the regions of high relevance within frames\footnote{The code can be found at \url{https://github.com/Benji4/zebrafish-learning.git}}.

\begin{figure}[t]
	\begin{center}
		\includegraphics[width=\columnwidth]{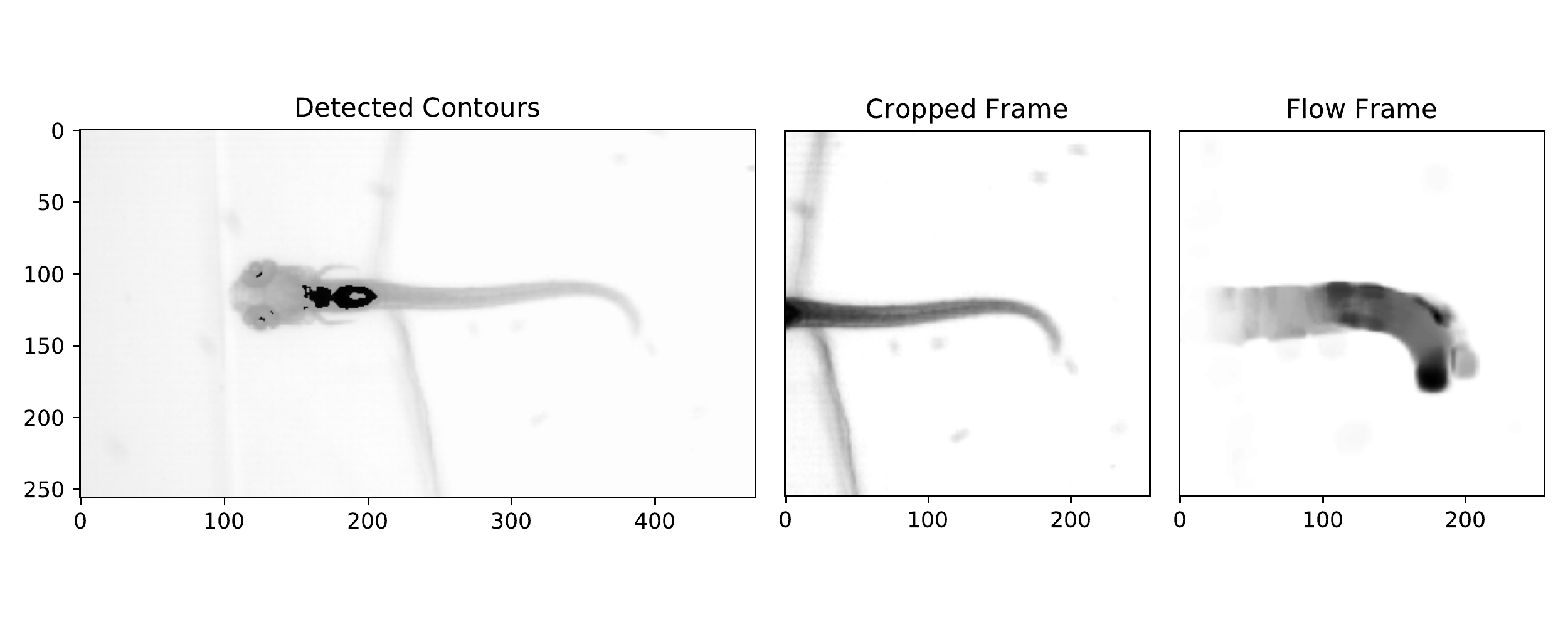}
		\caption{Detected contours (black) in the original frame, resulting cropped frame after normalization, and corresponding optical flow frame.}
		\label{fig:frame}
	\end{center}
\end{figure}

\textbf{Data pre-processing.}
We used the raw video files recorded by \cite{Semmelhack2014} with a high-speed camera at 300 frames per second and labeled as either spontaneous (0/negative, 56.1\%) or prey (1/positive, 43.9\%) bout.
The heads of the fish were embedded in a substance called agarose to keep them steady.
We turned the videos into grayscale, normalized them, and kept a crop of size 256x256 pixels from each frame, such that the right-most part of the bladder was central on the left, as shown in Figure~\ref{fig:frame}.
We did not include the head of the fish in the crop, because the eyes would give away information about the type of bout (\cite{Bianco2011}).
More details in Appendix~\ref{app:details}.

For correct centering we implemented a gamma correction with $\gamma = \exp{\left(\nicefrac{-\text{skewness}}{\text{param}}\right)}$, (where param was tweaked to maximize detection performance (here param~$= 4.3$)) and applied a binary threshold at value $3$ to separate the bladder from the rest of the fish.
Since the eyes might also fall under this threshold, we declared the right-most contour the bladder.
Each raw video contained several bout events which we extracted as sequences of 150 frames resulting in 1,214 video snippets.
Each sequence started 15 frames before the actual bout motion was detected.

\textbf{Extension to the pre-processing procedure.}
After training and heatmap analysis, we found that the trained CNN had found a shortcut for classification by using agarose motion instead of tail features.
We therefore extended our pre-processing procedure by setting squares of 85x85 pixels in the upper and lower left-hand corners of all frames plain white.
While this cut away some of the tail tip in rare occasions, it ensured that bouts could not be classified based on the agarose motion anymore.

\textbf{Data augmentation.}
Our data augmentation procedure worked with batches of 32 videos with the original data randomly shuffled, allowing a decent approximation of the gradient during training with some inherent noise, which is generally desirable to avoid falling into sharp minima (\cite{Keskar2016}).
The procedure performed three augmentation steps: first subsampling, then flipping, and cropping, which achieved augmentation factors of 8, 2, and 9 respectively, totaling 174,816 augmented samples.
We decided to compute optical flow after subsampling, not before, in order to create differing augmented samples.
All samples of an augmented batch were derived from different original videos to ensure a good gradient approximation during training.

In the subsampling step, our algorithm randomly selected 86 out of 150 frames under the constraint that between two frames, no more than 2 frames may be omitted.
This was to ensure meaningful flow calculation in the subsequent step, because the tail could move quite fast.
After subsampling, for each video our procedure selected one of the 86 frames to be the input for the spatial network.
Then we computed the optical flow resulting in 85 flow frames with one $x$ and $y$ component each.
They were stacked to 170 channels, alternating $x$ and $y$.
We used the optical flow algorithm by \cite{Farneback2003} with parameters\footnote{pyramid scale = 0.8, pyramid levels = 10, window size = 10, iterations per pyramid level = 10, pixel neighborhood for polynomial expansion = 13, std dev of the Gaussian for polynomial expansion = 1.8.} detecting flow even when the tail moved fast.
The procedure then generated 18 augmented batches from each subsampled batch by randomly flipping vertically and cropping into crops of size 224x224 with the upper left corner at coordinates (8,8), (8,16), (8,24), (16,8), (16,16), (16,24), (24,8), (24,16), and (24,24).

Since flow was calculated as floats, we minimized storage space significantly by rescaling each frame to range 0--255, compressing them with lossy JPEG-compression at level 40, and turning them into unsigned integers, similar to what \cite{Simonyan2014} had done.
Frames were rescaled later for training.

\begin{figure}[t]
	\begin{center}
		\includegraphics[width=\columnwidth]{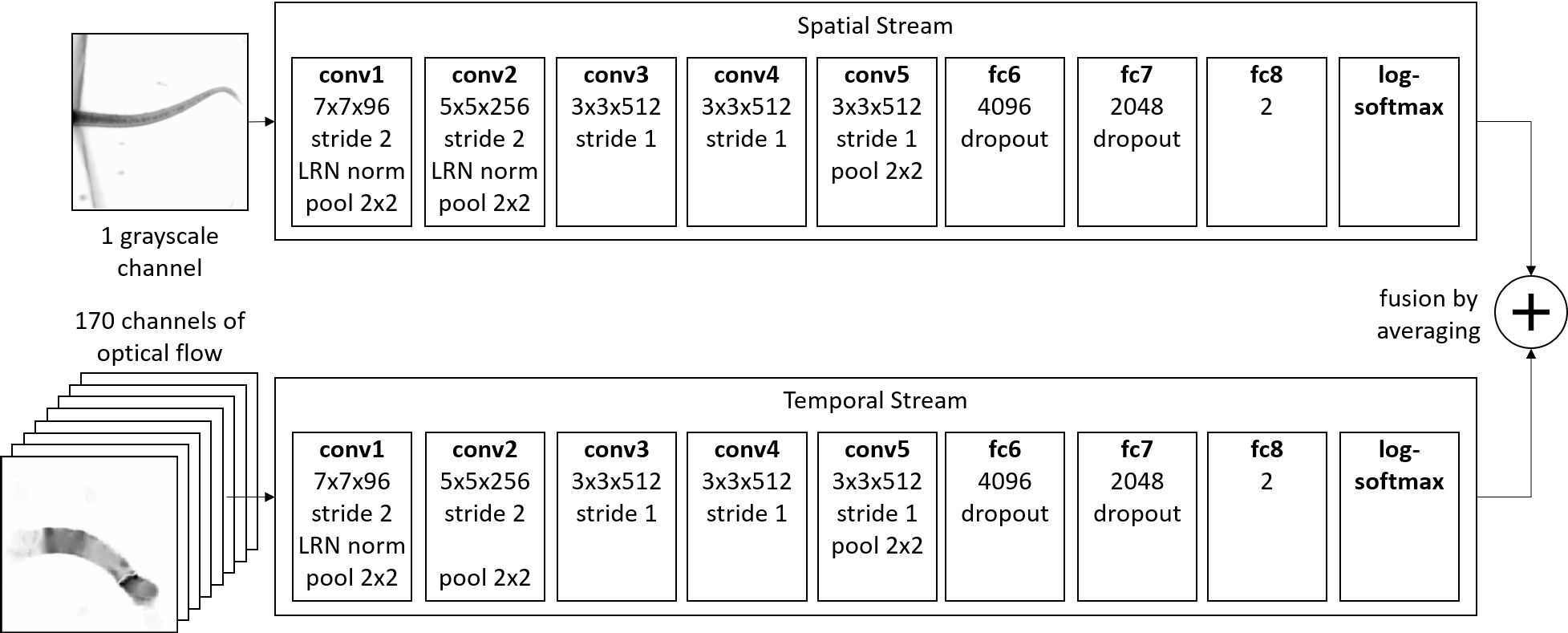}
		\caption{Full Two-Stream CNN architecture.}
		\label{fig:cnn}
	\end{center}
\end{figure}

\textbf{CNN architecture and framework.}
Just like \cite{Simonyan2014}, we used a two-stream network with an adapted CNN-M-2048 network (\cite{Chatfield2014}) for each stream, as depicted in Figure~\ref{fig:cnn}.
As shown by \cite{Carreira2017}, this network can deal with a small sample size and learns quickly.
The spatial stream had one gray-scale channel and the temporal stream 170 flow channels in the first layer.
After obtaining the predicted probabilities of each stream by calculating the log-softmax of the individual two outputs, they were fused by averaging.
We computed the negative log-likelihood loss of this average, which can be interpreted as the joint log-probability of both streams, assuming their statistical independence.

Our dataset was made up of 38 files, with 28, 4, and 6 files for training, validation, and test sets respectively.
% The next sentence is fuzzy.  You should elaborate. - Arno
The individual sets were not shuffled during training in order to allow sequential reads, which might have decreased training time.
Notwithstanding, batches consisted of random samples due to our augmentation procedure explained above.

\textbf{Initialization of weights.}
% I think I would move this to the main text. - Arno
We initialized both streams with weights pre-trained on ImageNet\footnote{http://www.vlfeat.org/matconvnet/models/imagenet-vgg-m-2048.mat}.
This has become a common initialization strategy in various classification tasks (\cite{Shin2016, Lee2016, VanHorn2017}), due to the utility of general features learnt from natural images, such as edge detection.
While \cite{Simonyan2014} did not pre-train flow, \cite{Carreira2017} found a pre-trained temporal stream to reach superior performance.
With this initialization we hoped that training would require only fine-tuning and therefore less data and fewer epochs.

Specifically, for the weights of the input layer of the spatial stream we took the average of the pre-trained weights of 3 RGB-channels to get the weights for 1 grayscale-channel.
For the temporal stream we copied the RGB-channels 56$\frac{2}{3}$ times to get 170 channels, and added uniform random noise to all of them.
This was to ensure that the channels evolved differently during training and should aid learning.
Regarding outputs, we averaged the pre-trained weights of 500 units on the output layer to obtain the weights of two output neurons, because we dealt with 2 classes instead of 1,000 as in ImageNet.

\textbf{Training procedure.}
We made use of the Adam optimizer with standard settings and tuned its learning rate and weight decay coefficient -- the neural network equivalent of L2-regularization (\cite{Loshchilov2019}).
Furthermore, we added a gamma learning rate scheduler which updated the learning rate by a multiplicative factor of $\gamma = \nicefrac{1}{\sqrt{epoch}}$ (\cite{Kingma2014}) every epoch.
Our training framework computed accuracy on the validation set after each epoch to estimate generalization performance.

Since we were initializing our CNN with weights learned from quite a different domain, fine-tuning was crucial.
In particular, we performed a hyperparameter search over learning rate (\{1e-3, 1e-4, 1e-5\}) and weight decay (\{1e-2, 1e-3, 1e-4\}) using a smaller dataset and training for about 8 epochs.
With the best hyperparameters we trained the CNN on the full dataset for 5 epochs.

\textbf{Relevance analysis with heatmaps.}
Making our CNN more transparent required an AI explainability technique which would be well interpretable in the optical flow domain of our temporal stream.
We expected feature visualization approaches (\cite{Olah2018, Bau2017, Erhan2009}) and deconvolutions (\cite{Zeiler2014}) to yield less interpretable outputs than attribution techniques.
In particular, we chose to analyze our CNN with Deep Taylor Decomposition (DTD), because it had been applied successfully before (\cite{Lapuschkin2019, VanMolle2018, Anders2018}) and was conveniently accessible in the ``iNNvestigate'' toolbox (\cite{Alber2018}).
For validation purposes, we additionally generated saliency maps (\cite{Simonyan2014b}) and heatmaps from Guided BackProp (\cite{Springenberg2015}).
The produced relevance heatmaps could be expected to give clues about what specific regions of optical flow, within a frame and across frames, the network was paying attention to.

Also, we simplified the analysis by splitting the network into its individual streams.
This was possible because no weights were learned after the final layer of each stream.
Once the network was initialized correctly, ``iNNvestigate'' made the generation of heatmaps surprisingly simple.
Also, with about 50 minutes for the whole analysis it was quite fast even on CPU, because it effectively only needed one forward and backward pass per sample.
We used a small dataset of 3,420 samples for analysis by setting the subsampling factor from 8 to 1, in order to simplify the process.

\section{Results} \label{sec:results}

% Summarize in the first sentence - or even paragraph - what you investigated and tried to achieve. - Arno
We fine-tuned the training of our CNN to reach high accuracies in distinguishing prey bouts of larval zebrafish from spontaneous swims.
Subsequently, we analyzed the learned weights by computing and averaging the relevance heatmaps of all samples grouped by class.
We further define prey bouts as positive and spontaneous swims as negative.

\textbf{CNN test results.}
We performed a small hyperparameter search over learning rate and weight decay, which proved sufficient because all models were initialized with pre-trained weights from ImageNet.
For our baseline SVM -- detailed in Appendix~\ref{app:svm} --
% Insert reference to appendix. - Arno
we report the 5-fold cross-validated accuracy and the final accuracy on the held-out test set.
The hyperparameters\footnote{RBF-kernel with $\gamma$ = 0.001 and C = 1} agree with the ones found by \cite{Semmelhack2014}.
We further present the accuracies of the CNN used for heatmap analysis, as well as the best CNNs before and after removal of experimental artifacts on training, validation, and test sets in Table~\ref{tab:exps}.
For the test set we additionally report individual accuracies of the spatial and temporal streams.
We highlight that the final CNN attains a test accuracy of 96.32\%, which is 6.12\% points better than the baseline.

\begin{table}[t]
	\caption{Results on training, validation, and test (split by stream) sets for baseline (B), analyzed CNN (0), best CNN before (1) and after (2) correction.}
	\label{tab:exps}
	\begin{center}
		\begin{tabular}{cccccccc}
			\multicolumn{1}{c}{\bf ID}  &\multicolumn{1}{c}{\bf LR} &\multicolumn{1}{c}{\bf WD} &\multicolumn{1}{c}{\bf TRAIN} &\multicolumn{1}{c}{\bf VALID} &\multicolumn{1}{c}{\bf SPATIAL} &\multicolumn{1}{c}{\bf TEMPORAL}
			&\multicolumn{1}{c}{\bf FULL TEST}
			\\ \hline \\
			B & -- & -- & .946 & .946 & -- & -- & .902 \\
			0  & 1e-4 & 1e-3 & .9962 & .9489 & .8216  & .9441 & .9596 \\
			1  & 1e-5 & 1e-3 & .9869 & .9372 & .8138  & .9660 & .9669 \\
			2  & 1e-5 & 1e-3 & .9997 & .9597 & .8213  & .9581 & .9632 \\
		\end{tabular}
	\end{center}
\end{table}

\begin{figure}[t]
	\centering
	\begin{subfigure}{.195\textwidth}
		\centering
		\includegraphics[width=\linewidth]{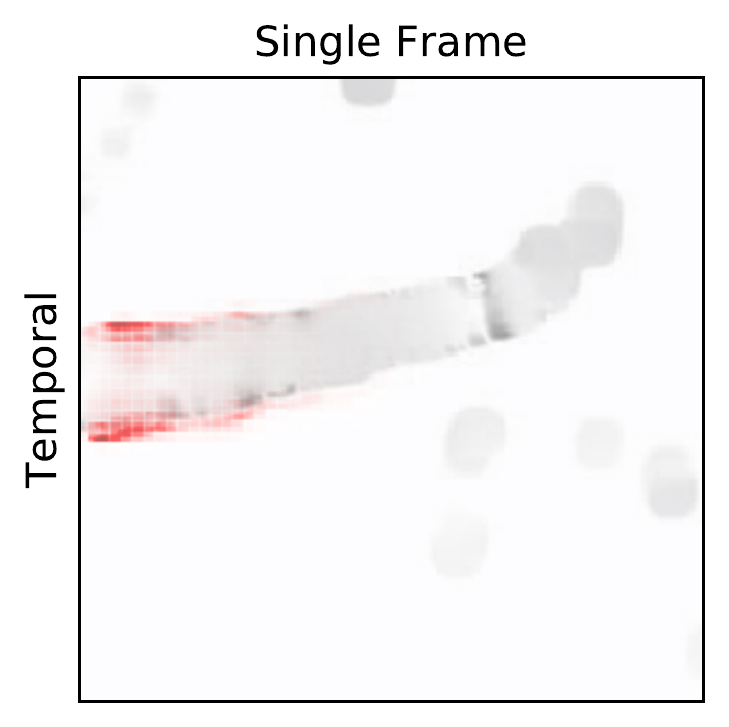}
	\end{subfigure}
	\begin{subfigure}{.55\textwidth}
		\centering
		\includegraphics[width=\linewidth]{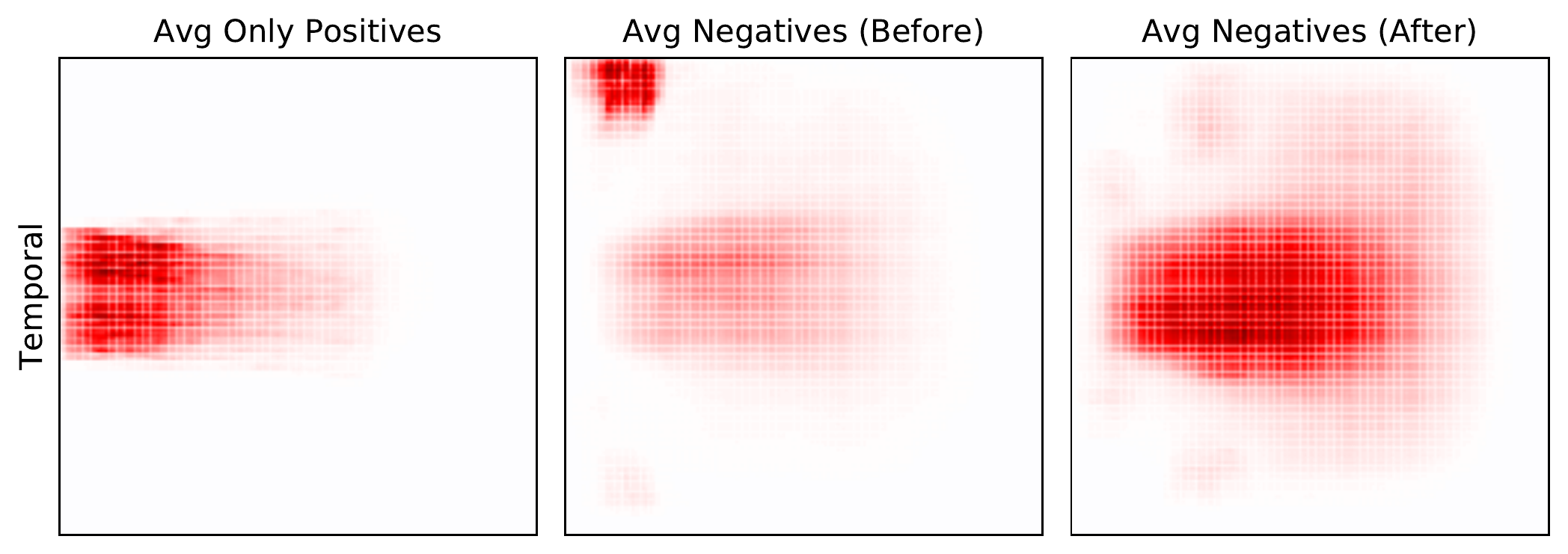}
	\end{subfigure}
	\begin{subfigure}{.22\textwidth}
		\vskip 5pt
		\centering
		\includegraphics[width=\linewidth]{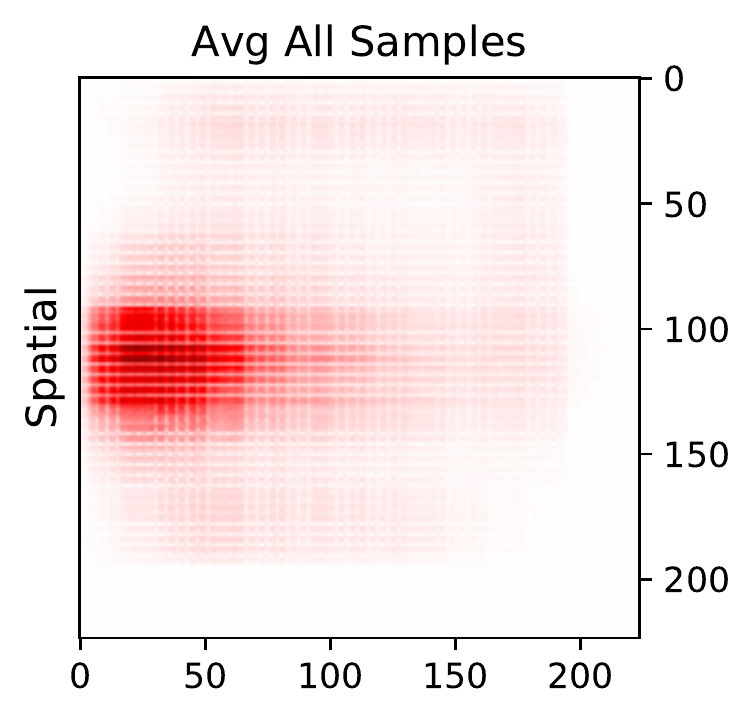}
	\end{subfigure}
	\caption{
		Relevance heatmaps of the CNN. High relevance is shown in dark red color, while light red stands for low relevance. Left: example of a temporal heatmap over a single frame (true positive, optical flow in grayscale, relevance heatmap overlaid in red). Middle: averages of temporal heatmaps split by class; negatives before and after correction. Right: average of spatial heatmaps over all samples.}
	\label{fig:the_average_frame}
\end{figure}

\textbf{Relevance heatmaps.}
We used relevance heatmaps to visualize the regions the CNN pays most attention to when classifying a specific sample.
We computed relevance averages across samples and frames, as well as split by class in Figure~\ref{fig:the_average_frame} for more comprehensive insights into the features learned by our CNN.
Similar results from other explainability techniques can be found in Appendix~\ref{app:sens}.
As expected, the heatmaps exhibit the checkerboard artifacts typical of kernels with stride~2 in the first convolutional layer (\cite{Odena2016, Montavon2017}).

\textbf{Steadiness of the fish's trunk as differentiating feature.}
% Results - Arno
First and foremost, the heatmaps show that our CNN is able to differentiate the movements of zebrafish based on their characteristic motion.
Relevance is highly concentrated at the trunk, i.e. the rostral part of the tail, for both temporal and spatial stream.
We observe a very sharp relevance pattern along the edges of the tail.
This indicates that the pre-trained weights helped the network to look for edges.
The CNN pays little to no attention to the end of the tail even though it is visible in most frames.
Instead, it makes positive classifications by looking at the very start of the trunk.
The heatmaps suggest that a calm and steady trunk indicates a prey bout.
As for the negatives, the spread out relevance pattern reflects the high frequency of tail deflections typical of spontaneous bouts, which had been identified by \cite{Semmelhack2014} before.
This makes clear that the network is able to differentiate the movements of zebrafish to high accuracy based on their characteristic motion.

\textbf{``Clever Hans'' predictions.}
% Results - Arno
CNNs are incredibly powerful at finding any kinds of correlations in the input data even if they are not related to the object of interest.
\cite{Lapuschkin2019} have termed such spurious correlations ``Clever Hans'' predictions, because the model bases its prediction not on what we want it to focus on, but some unintended artifacts in the data.
Figure~\ref{fig:the_average_frame} shows clearly that our CNN bases a significant number of its negative responses mainly on motion in the top left corner and focuses little on tail appearance and motion.
This plays a role only in negative classifications and only in the temporal stream.
While the heatmaps are vertically symmetric, as we would expect due to vertical flipping during augmentation, this is not true for the peculiar region in the top left corner.
Figure~\ref{fig:the_average_frame} depicts the averaged heatmap after removing the artifacts in the top and bottom left hand corners and retraining our CNN.
Relevance is now entirely focused on the tail.

\textbf{Relevance distribution across frames.}
\begin{figure}[t]
	\begin{center}
		\includegraphics[width=\columnwidth]{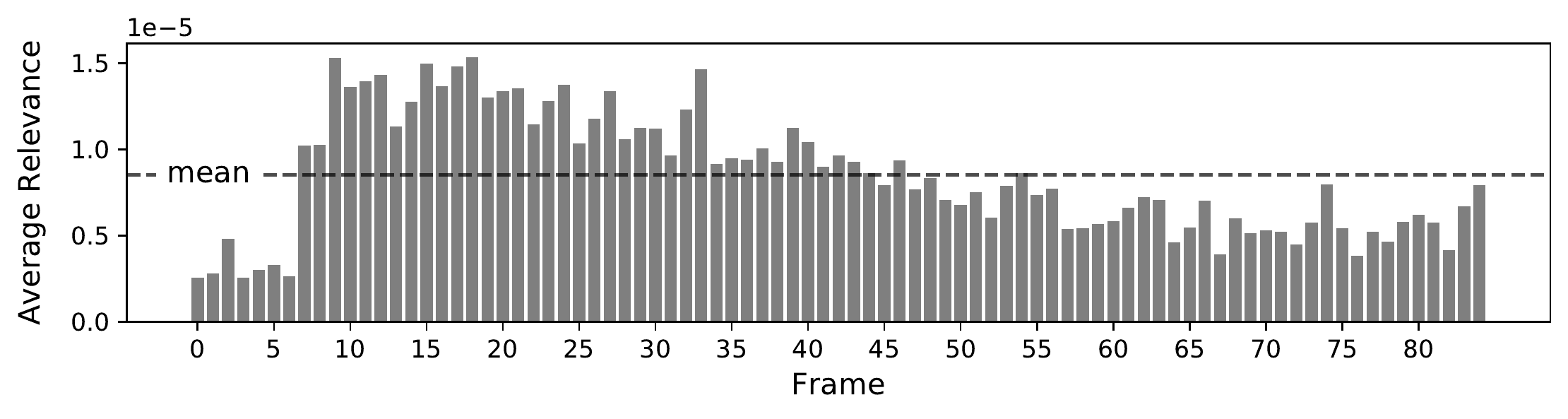}
		\caption{Distribution of relevance on the flow frames of the averaged sample.}
		\label{fig:bars}
	\end{center}
\end{figure}
Most relevance is concentrated on the frames in the range 7--46, as depicted in Figure~\ref{fig:bars}.
The first seven frames are of least importance.
This is very likely because our pre-processing procedure added a buffer of 15 frames before each bout, suggesting that the network focuses on the range of frames which are in fact the most relevant ones.
This further supports the hypothesis that our CNN is able to differentiate zebrafish movements based on their characteristic motion patterns.

\section{Conclusion and Future Work} \label{sec:conclusion}

We trained a two-stream Convolutional Neural Network (CNN) on recordings of larval zebrafish to classify prey and spontaneous swim bouts.
We then visualized the learned weights by generating relevance heatmaps showing which regions of the input the network focuses on while performing its classifications.
We find that our CNN is capable of learning highly discriminating tail features.
These features seem to be quite different from the ones used in the SVM classification by \cite{Semmelhack2014} - the previous state-of-the-art in bout classification.
The heatmaps further uncovered a ``Clever Hans'' type of correlation.
% You should emphasize the next sentence also in results. - Arno
After removing this spurious correlation and re-training the network, the network reached a test accuracy of 96.32\%, which is 6.12\% points better than the accuracy achieved by \cite{Semmelhack2014}.
Judging from the test accuracy, our CNN has learned better discriminating features than those used for the SVM by \cite{Semmelhack2014}, and has thus beaten manual feature engineering in this application domain.

\textbf{Steadiness of the fish's trunk as differentiating feature.}
The relevance heatmaps and high accuracy show that the network achieves correct classifications by looking for salient features in the trunk of the tail while largely disregarding the tip.
A sharp and clear relevance profile confined to the edges of the trunk gives a clear sign of a prey bout.
The opposite speaks for a spontaneous bout.
Here, attention spreads out to capture the strong vertical oscillation of the trunk.
For this reason we conclude that the CNN makes its predictions based on the steadiness of the trunk.
We believe our interpretation of learned features to be in line with existing research on the kinematics of prey bouts.
As shown by \cite{Borla2002} and \cite{McElligott2005}, prey bouts require fine control of the tail's axial kinematics to perform precise swim movements.
Zebrafish noticeably reduce their yaw rotation and stabilize the positioning of their head to make a targeted move at their prey.
Such precise movements are not required in spontaneous swim bouts.
The heatmaps indicate that the network has found clear evidence for these kinds of motion in the trunk of the tail.

% Next paragraph -> Conclusions - Arno
Furthermore, we argue that the CNN has learned features which are very different from the ones identified by \cite{Semmelhack2014}.
All of their features --~as outlined in Section~\ref{sec:related}~--, except the second one, rely on information from the tip of the tail and a complete sequence of frames.
However, many optical flow frames do not depict the tip of the tail because of its small size and high speed.
This might have happened due to suboptimal parameter settings which could not handle the sometimes long distances which the tip traveled between frames.
Also, subsamples include only 85 of the original 150 frames for each video.
Due to its higher performance, we conclude not only that the CNN has learned a different set of features, but also that these features must bear higher discriminative power.

% Conclusions - Arno
\textbf{Origin of the ``Clever Hans'' correlation.}
The telltale motion in the top left corner stems from a substance called agarose, which the fish's head was embedded in to keep it steady.
It is quite curious that, while not visible to human eyes, the agarose seems to be moving each time the fish performed a spontaneous swim bout, but not so for a prey bout.
We speculate that this correlation was unintentionally introduced by the experimenters who might have tapped the petri dish to induce the fish to perform a spontaneous swim bout.

% This future work paragraph is ok but given the target audience, a technical paragraph on potential computational method innovation would be preferable. - Arno

\textbf{Future work.}
Calculating and storing optical flow is expensive. If we attained similar performance on original frames, training would be considerably cheaper.
While we can confirm the findings by \cite{Simonyan2014} that the spatial stream by itself reaches a fairly competitive accuracy, it provides only very minor improvement to the overall network.
Yet, this stream is probably looking for very similar features as the temporal stream, because it focuses largely on the upper half of the tail, just like the temporal stream.
If that is the case, we should see improved performance when giving the spatial stream a sequence of frames. 
It should be interesting to probe whether the spatial stream could then match or even surpass the performance of the temporal stream.

Furthermore, CNNs such as the one used in this paper could be used to investigate brain recovery in larval zebrafish.
It has been shown on a cellular level that zebrafish can heal their brain within days after a lesion.
However, this needs to be proven on a behavioral level (\cite{Krakauer2017}).
Future work could perform a lesion study on the optic tectum in zebrafish (\cite{McDowell2004, Roeser2003}), a brain region responsible for translating visual input into motor output.
CNNs could then assess swim bouts of recovered fish and give a measure for potential behavioral changes.
Insights from relevance heatmaps would be required if the CNN were not able not distinguish recovered fish from healthy ones.

%Another interesting project would be comparing a partially restrained experimental setup, where the fish is embedded in agarose, with the natural setup of freely moving fish.
%Fixated fish are easier to observe, for example for live brain imaging \cite{Kim2017}, and existing research is based on the assumption that they move the same as in a free setup.
%CNNs can be used to challenge this assumption by comparing swim bouts of each setup and even give hints as to what might be discriminating features.

%\subsubsection*{Author Contributions}
%If you'd like to, you may include  a section for author contributions as is done
%in many journals. This is optional and at the discretion of the authors.
%
%\subsubsection*{Acknowledgments}
%Use unnumbered third level headings for the acknowledgments. All
%acknowledgments, including those to funding agencies, go at the end of the paper.

\bibliography{iclr2020_conference}

\begin{thebibliography}{61}
\providecommand{\natexlab}[1]{#1}
\providecommand{\url}[1]{\texttt{#1}}
\expandafter\ifx\csname urlstyle\endcsname\relax
  \providecommand{\doi}[1]{doi: #1}\else
  \providecommand{\doi}{doi: \begingroup \urlstyle{rm}\Url}\fi

\bibitem[Abadi et~al.(2015)Abadi, Agarwal, Barham, Brevdo, Chen, Citro,
  Corrado, Davis, Dean, Devin, Ghemawat, Goodfellow, Harp, Irving, and
  Others]{tensorflow2015}
Mart\'{\i}n Abadi, Ashish Agarwal, Paul Barham, Eugene Brevdo, Zhifeng Chen,
  Craig Citro, Greg~S. Corrado, Andy Davis, Jeffrey Dean, Matthieu Devin,
  Sanjay Ghemawat, Ian Goodfellow, Andrew Harp, Geoffrey Irving, and Others.
\newblock {TensorFlow}: Large-scale machine learning on heterogeneous systems,
  2015.
\newblock URL \url{http://tensorflow.org/}.
\newblock Software available from tensorflow.org.

\bibitem[Alber et~al.(2018)Alber, Lapuschkin, Seegerer, H{\"{a}}gele,
  Sch{\"{u}}tt, Montavon, Samek, M{\"{u}}ller, D{\"{a}}hne, and
  Kindermans]{Alber2018}
Maximilian Alber, Sebastian Lapuschkin, Philipp Seegerer, Miriam H{\"{a}}gele,
  Kristof~T. Sch{\"{u}}tt, Gr{\'{e}}goire Montavon, Wojciech Samek,
  Klaus-Robert M{\"{u}}ller, Sven D{\"{a}}hne, and Pieter-Jan Kindermans.
\newblock {iNNvestigate neural networks!}
\newblock aug 2018.
\newblock URL \url{http://arxiv.org/abs/1808.04260}.

\bibitem[Anders et~al.(2018)Anders, Montavon, Samek, and
  M{\"{u}}ller]{Anders2018}
Christopher Anders, Gr{\'{e}}goire Montavon, Wojciech Samek, and Klaus-Robert
  M{\"{u}}ller.
\newblock {Understanding Patch-Based Learning by Explaining Predictions}.
\newblock jun 2018.
\newblock URL \url{http://arxiv.org/abs/1806.06926}.

\bibitem[Arras et~al.(2017)Arras, Horn, Montavon, M{\"{u}}ller, and
  Samek]{Arras2017}
Leila Arras, Franziska Horn, Gr{\'{e}}goire Montavon, Klaus~Robert
  M{\"{u}}ller, and Wojciech Samek.
\newblock {"What is relevant in a text document?": An interpretable machine
  learning approach}.
\newblock \emph{PLoS ONE}, 12\penalty0 (8):\penalty0 e0181142, aug 2017.
\newblock URL \url{https://dx.plos.org/10.1371/journal.pone.0181142}.

\bibitem[Bach et~al.(2015)Bach, Binder, Montavon, Klauschen, M{\"{u}}ller, and
  Samek]{Bach2015}
Sebastian Bach, Alexander Binder, Gr{\'{e}}goire Montavon, Frederick Klauschen,
  Klaus~Robert M{\"{u}}ller, and Wojciech Samek.
\newblock {On pixel-wise explanations for non-linear classifier decisions by
  layer-wise relevance propagation}.
\newblock \emph{PLoS ONE}, 10\penalty0 (7):\penalty0 e0130140, jul 2015.
\newblock URL \url{https://dx.plos.org/10.1371/journal.pone.0130140}.

\bibitem[Bau et~al.(2017)Bau, Zhou, Khosla, Oliva, and Torralba]{Bau2017}
David Bau, Bolei Zhou, Aditya Khosla, Aude Oliva, and Antonio Torralba.
\newblock {Network Dissection: Quantifying Interpretability of Deep Visual
  Representations}, 2017.
\newblock URL
  \url{http://openaccess.thecvf.com/content_cvpr_2017/html/Bau_Network_Dissection_Quantifying_CVPR_2017_paper.html}.

\bibitem[Becker et~al.(2018)Becker, Ackermann, Lapuschkin, M{\"{u}}ller, and
  Samek]{Becker2018}
S{\"{o}}ren Becker, Marcel Ackermann, Sebastian Lapuschkin, Klaus-Robert
  M{\"{u}}ller, and Wojciech Samek.
\newblock {Interpreting and Explaining Deep Neural Networks for Classification
  of Audio Signals}.
\newblock jul 2018.
\newblock URL \url{http://arxiv.org/abs/1807.03418}.

\bibitem[Bianco et~al.(2011)Bianco, Kampff, and Engert]{Bianco2011}
Isaac~H. Bianco, Adam~R. Kampff, and Florian Engert.
\newblock {Prey Capture Behavior Evoked by Simple Visual Stimuli in Larval
  Zebrafish}.
\newblock \emph{Frontiers in Systems Neuroscience}, 5:\penalty0 101, dec 2011.
\newblock URL
  \url{http://journal.frontiersin.org/article/10.3389/fnsys.2011.00101/abstract}.

\bibitem[Borla et~al.(2002)Borla, Palecek, Budick, and O'Malley]{Borla2002}
Melissa~A Borla, Betsy Palecek, Seth Budick, and Donald~M. O'Malley.
\newblock {Prey capture by larval zebrafish: Evidence for fine axial motor
  control}.
\newblock \emph{Brain, Behavior and Evolution}, 60\penalty0 (4):\penalty0
  207--229, 2002.
\newblock URL \url{www.karger.comwww.karger.com/bbe}.

\bibitem[Bradski(2000)]{opencv_library}
G.~Bradski.
\newblock {The OpenCV Library}.
\newblock \emph{Dr. Dobb's Journal of Software Tools}, 2000.

\bibitem[Brox et~al.(2004)Brox, Bruhn, Papenberg, and Weickert]{Brox2004}
Thomas Brox, Andr{\'{e}}s Bruhn, Nils Papenberg, and Joachim Weickert.
\newblock {High Accuracy Optical Flow Estimation Based on a Theory for
  Warping}.
\newblock pp.\  25--36. Springer, Berlin, Heidelberg, 2004.
\newblock URL \url{http://link.springer.com/10.1007/978-3-540-24673-2_3}.

\bibitem[Carreira \& Zisserman(2017)Carreira and Zisserman]{Carreira2017}
Jo{\~{a}}o Carreira and Andrew Zisserman.
\newblock {Quo Vadis, action recognition? A new model and the kinetics
  dataset}.
\newblock In \emph{Proceedings - 30th IEEE Conference on Computer Vision and
  Pattern Recognition, CVPR 2017}, volume 2017-Janua, pp.\  4724--4733, 2017.
\newblock URL \url{https://arxiv.org/pdf/1705.07750.pdf}.

\bibitem[Chatfield et~al.(2014)Chatfield, Simonyan, Vedaldi, and
  Zisserman]{Chatfield2014}
Ken Chatfield, Karen Simonyan, Andrea Vedaldi, and Andrew Zisserman.
\newblock {Return of the Devil in the Details: Delving Deep into Convolutional
  Nets}.
\newblock 2014.
\newblock URL \url{http://www.robots.ox.ac.uk/ http://arxiv.org/abs/1405.3531}.

\bibitem[Chollet et~al.(2015)]{keras}
Fran\c{c}ois Chollet et~al.
\newblock Keras.
\newblock \url{https://github.com/fchollet/keras}, 2015.

\bibitem[Collette(2013)]{hdf5}
Andrew Collette.
\newblock \emph{Python and HDF5}.
\newblock O'Reilly, 2013.

\bibitem[da~Costa-Luis(2019)]{DaCosta-Luis2019}
Casper~O. da~Costa-Luis.
\newblock {tqdm: A Fast, Extensible Progress Meter for Python and CLI}.
\newblock \emph{Journal of Open Source Software}, 4\penalty0 (37):\penalty0
  1277, 2019.
\newblock \doi{10.21105/joss.01277}.
\newblock URL \url{https://hub.docker.com/}.

\bibitem[Dosovitskiy et~al.(2015)Dosovitskiy, Fischer, Ilg, Hausser, Hazirbas,
  Golkov, van~der Smagt, Cremers, and Brox]{Dosovitskiy2015}
Alexey Dosovitskiy, Philipp Fischer, Eddy Ilg, Philip Hausser, Caner Hazirbas,
  Vladimir Golkov, Patrick van~der Smagt, Daniel Cremers, and Thomas Brox.
\newblock {FlowNet: Learning Optical Flow With Convolutional Networks}, 2015.
\newblock URL
  \url{http://openaccess.thecvf.com/content_iccv_2015/html/Dosovitskiy_FlowNet_Learning_Optical_ICCV_2015_paper.html}.

\bibitem[Erhan et~al.(2009)Erhan, Courville, Bengio, and Vincent]{Erhan2009}
Dumitru Erhan, Aaron Courville, Yoshua Bengio, and Pascal Vincent.
\newblock {Visualizing Higher-Layer Features of a Deep Network Oracle
  Performance for Visual Captioning View project Semantic View project
  Visualizing Higher-Layer Features of a Deep Network D{\'{e}}partement
  d'Informatique et Recherche Op{\'{e}}rationnelle}.
\newblock \penalty0 (August 2014), 2009.
\newblock URL \url{https://www.researchgate.net/publication/265022827}.

\bibitem[Farneb{\"{a}}ck(2003)]{Farneback2003}
Gunnar Farneb{\"{a}}ck.
\newblock {Two-Frame Motion Estimation Based on Polynomial Expansion}.
\newblock pp.\  363--370. 2003.
\newblock URL \url{http://www.isy.liu.se/cvl/}.

\bibitem[Gahtan(2005)]{Gahtan2005}
Ethan Gahtan.
\newblock {Visual Prey Capture in Larval Zebrafish Is Controlled by Identified
  Reticulospinal Neurons Downstream of the Tectum}.
\newblock \emph{Journal of Neuroscience}, 25\penalty0 (40):\penalty0
  9294--9303, 2005.
\newblock URL \url{http://rsb.info.nih.gov/ij/
  http://www.jneurosci.org/cgi/doi/10.1523/JNEUROSCI.2678-05.2005}.

\bibitem[Hinton et~al.(2012)Hinton, Srivastava, Krizhevsky, Sutskever, and
  Salakhutdinov]{Hinton2012}
Geoffrey~E. Hinton, Nitish Srivastava, Alex Krizhevsky, Ilya Sutskever, and
  Ruslan~R. Salakhutdinov.
\newblock {Improving neural networks by preventing co-adaptation of feature
  detectors}.
\newblock jul 2012.
\newblock URL \url{http://arxiv.org/abs/1207.0580}.

\bibitem[Howe et~al.(2013)Howe, Clark, Torroja, Torrance, Berthelot, Muffato,
  Collins, Humphray, McLaren, Matthews, and Others]{Howe2013}
Kerstin Howe, Matthew~D Clark, Carlos~F Torroja, James Torrance, Camille
  Berthelot, Matthieu Muffato, John~E Collins, Sean Humphray, Karen McLaren,
  Lucy Matthews, and And Others.
\newblock {The zebrafish reference genome sequence and its relationship to the
  human genome}.
\newblock \emph{Nature}, 496\penalty0 (7446):\penalty0 498--503, apr 2013.
\newblock URL \url{http://www.nature.com/doifinder/10.1038/nature12111}.

\bibitem[Hunter(2007)]{Hunter2007}
John~D. Hunter.
\newblock {Matplotlib: A 2D Graphics Environment}.
\newblock \emph{Computing in Science {\&} Engineering}, 9\penalty0
  (3):\penalty0 90--95, 2007.
\newblock URL \url{http://ieeexplore.ieee.org/document/4160265/}.

\bibitem[Hwang et~al.(2013)Hwang, Fu, Reyon, Maeder, Tsai, Sander, Peterson,
  Yeh, and Joung]{Hwang2013}
Woong~Y Hwang, Yanfang Fu, Deepak Reyon, Morgan~L Maeder, Shengdar~Q Tsai,
  Jeffry~D Sander, Randall~T Peterson, J-R~Joanna Yeh, and J~Keith Joung.
\newblock {Efficient genome editing in zebrafish using a CRISPR-Cas system.}
\newblock \emph{Nature biotechnology}, 31\penalty0 (3):\penalty0 227--9, 2013.
\newblock URL \url{http://www.nature.com/doifinder/10.1038/nbt.2501.}

\bibitem[Ilg et~al.(2017)Ilg, Mayer, Saikia, Keuper, Dosovitskiy, and
  Brox]{Ilg2017}
Eddy Ilg, Nikolaus Mayer, Tonmoy Saikia, Margret Keuper, Alexey Dosovitskiy,
  and Thomas Brox.
\newblock {FlowNet 2.0: Evolution of Optical Flow Estimation With Deep
  Networks}, 2017.
\newblock URL
  \url{http://openaccess.thecvf.com/content_cvpr_2017/html/Ilg_FlowNet_2.html}.

\bibitem[Ji et~al.(2013)Ji, Xu, Yang, and Yu]{Ji2013}
Shuiwang Ji, Wei Xu, Ming Yang, and Kai Yu.
\newblock {3D Convolutional neural networks for human action recognition}.
\newblock \emph{IEEE Transactions on Pattern Analysis and Machine
  Intelligence}, 35\penalty0 (1):\penalty0 221--231, 2013.

\bibitem[Keskar et~al.(2016)Keskar, Mudigere, Nocedal, Smelyanskiy, and
  Tang]{Keskar2016}
Nitish~Shirish Keskar, Dheevatsa Mudigere, Jorge Nocedal, Mikhail Smelyanskiy,
  and Ping Tak~Peter Tang.
\newblock {On Large-Batch Training for Deep Learning: Generalization Gap and
  Sharp Minima}.
\newblock sep 2016.
\newblock URL \url{http://arxiv.org/abs/1609.04836}.

\bibitem[Kingma \& Ba(2014)Kingma and Ba]{Kingma2014}
Diederik~P. Kingma and Jimmy Ba.
\newblock {Adam: A Method for Stochastic Optimization}.
\newblock dec 2014.
\newblock URL \url{http://arxiv.org/abs/1412.6980}.

\bibitem[Kishimoto et~al.(2011)Kishimoto, Shimizu, and Sawamoto]{Kishimoto2011}
N.~Kishimoto, K.~Shimizu, and K.~Sawamoto.
\newblock {Neuronal regeneration in a zebrafish model of adult brain injury}.
\newblock \emph{Disease Models {\&} Mechanisms}, 5\penalty0 (2):\penalty0
  200--209, 2011.
\newblock URL \url{http://dmm.biologists.org/content/dmm/5/2/200.full.pdf}.

\bibitem[Kizil et~al.(2012)Kizil, Kaslin, Kroehne, and Brand]{Kizil2012}
Caghan Kizil, Jan Kaslin, Volker Kroehne, and Michael Brand.
\newblock {Adult neurogenesis and brain regeneration in zebrafish}.
\newblock \emph{Developmental Neurobiology}, 72\penalty0 (3):\penalty0
  429--461, 2012.
\newblock URL \url{https://onlinelibrary.wiley.com/doi/pdf/10.1002/dneu.20918}.

\bibitem[Kohavi(1995)]{Kohavi1995}
Ron Kohavi.
\newblock {A Study of Cross-Validation and Bootstrap for Accuracy Estimation
  and Model Selection}.
\newblock \emph{International Joint Conference of Artificial Intelligence},
  1995.
\newblock URL \url{http://robotics.stanford.edu/ronnyk}.

\bibitem[Krakauer et~al.(2017)Krakauer, Ghazanfar, Gomez-Marin, MacIver, and
  Poeppel]{Krakauer2017}
John~W. Krakauer, Asif~A. Ghazanfar, Alex Gomez-Marin, Malcolm~A. MacIver, and
  David Poeppel.
\newblock {Neuroscience Needs Behavior: Correcting a Reductionist Bias}, feb
  2017.
\newblock URL
  \url{https://www.sciencedirect.com/science/article/pii/S0896627316310406}.

\bibitem[Krizhevsky et~al.(2012)Krizhevsky, Sutskever, and
  Hinton]{Krizhevsky2012}
Alex Krizhevsky, Ilya Sutskever, and Geoffrey~E Hinton.
\newblock {Imagenet classification with deep convolutional neural networks}.
\newblock \emph{Advances In Neural Information Processing Systems}, pp.\  1--9,
  2012.
\newblock URL
  \url{https://papers.nips.cc/paper/4824-imagenet-classification-with-deep-convolutional-neural-networks.pdf}.

\bibitem[Lapuschkin et~al.(2019)Lapuschkin, W{\"{a}}ldchen, Binder, Montavon,
  Samek, and M{\"{u}}ller]{Lapuschkin2019}
Sebastian Lapuschkin, Stephan W{\"{a}}ldchen, Alexander Binder, Gr{\'{e}}goire
  Montavon, Wojciech Samek, and Klaus~Robert M{\"{u}}ller.
\newblock {Unmasking Clever Hans predictors and assessing what machines really
  learn}.
\newblock \emph{Nature Communications}, 10\penalty0 (1):\penalty0 1096, dec
  2019.
\newblock URL \url{http://www.nature.com/articles/s41467-019-08987-4}.

\bibitem[Lecun \& Bengio(1995)Lecun and Bengio]{Lecun1995}
Y~Lecun and Y~Bengio.
\newblock {Convolutional networks for images, speech, and time-series}.
\newblock Technical report, 1995.
\newblock URL \url{https://www.researchgate.net/publication/2453996}.

\bibitem[Lee et~al.(2016)Lee, Chang, Chan, and Remagnino]{Lee2016}
Sue~Han Lee, Yang~Loong Chang, Chee~Seng Chan, and Paolo Remagnino.
\newblock {Plant identification system based on a convolutional neural network
  for the lifeclef 2016 plant classification task}.
\newblock In \emph{CEUR Workshop Proceedings}, volume 1609, pp.\  502--510,
  2016.

\bibitem[Leslie(2019)]{Leslie2019}
David Leslie.
\newblock \emph{{Understanding artificial intelligence ethics and safety: A
  guide for the responsible design and implementation of AI systems in the
  public sector.}}
\newblock 2019.
\newblock URL \url{https://doi.org/10.5281/zenodo.3240529}.

\bibitem[Loshchilov \& Hutter(2019)Loshchilov and Hutter]{Loshchilov2019}
Ilya Loshchilov and Frank Hutter.
\newblock {Fixing Weight Decay Regularization in Adam}.
\newblock In \emph{Proceedings of the 2019 International Conference on Learning
  Representations (ICLR'19)}, 2019.
\newblock URL \url{http://arxiv.org/abs/1711.05101}.

\bibitem[McDowell et~al.(2004)McDowell, Dixon, Houchins, and
  Bilotta]{McDowell2004}
Angela~L. McDowell, Lee~J Dixon, Jennifer~D Houchins, and Joseph Bilotta.
\newblock {Visual processing of the zebrafish optic tectum before and after
  optic nerve damage}.
\newblock \emph{Visual Neuroscience}, 21\penalty0 (2):\penalty0 97--106, 2004.
\newblock URL \url{https://doi.org/10.1017/S0952523804043019}.

\bibitem[McElligott \& O'Malley(2005)McElligott and O'Malley]{McElligott2005}
Melissa~B. McElligott and Donald~M. O'Malley.
\newblock {Prey tracking by larval zebrafish: Axial kinematics and visual
  control}.
\newblock \emph{Brain, Behavior and Evolution}, 66\penalty0 (3):\penalty0
  177--196, 2005.
\newblock URL \url{www.karger.com}.

\bibitem[Molnar(2019)]{Molnar2019}
Christoph Molnar.
\newblock {Interpretable Machine Learning}, 2019.
\newblock URL \url{https://christophm.github.io/interpretable-ml-book/}.

\bibitem[Montavon et~al.(2017)Montavon, Lapuschkin, Binder, Samek, and
  M{\"{u}}ller]{Montavon2017}
Gr{\'{e}}goire Montavon, Sebastian Lapuschkin, Alexander Binder, Wojciech
  Samek, and Klaus~Robert M{\"{u}}ller.
\newblock {Explaining nonlinear classification decisions with deep Taylor
  decomposition}.
\newblock \emph{Pattern Recognition}, 65:\penalty0 211--222, may 2017.
\newblock URL
  \url{https://www.sciencedirect.com/science/article/pii/S0031320316303582}.

\bibitem[Odena et~al.(2016)Odena, Dumoulin, and Olah]{Odena2016}
Augustus Odena, Vincent Dumoulin, and Chris Olah.
\newblock {Deconvolution and Checkerboard Artifacts}.
\newblock \emph{Distill}, 1\penalty0 (10):\penalty0 e3, oct 2016.
\newblock URL \url{http://distill.pub/2016/deconv-checkerboard}.

\bibitem[Olah et~al.(2017)Olah, Mordvintsev, and Schubert]{Olah2017}
Chris Olah, Alexander Mordvintsev, and Ludwig Schubert.
\newblock {Feature Visualization}.
\newblock \emph{Distill}, 2\penalty0 (11):\penalty0 e7, nov 2017.
\newblock URL \url{https://distill.pub/2017/feature-visualization}.

\bibitem[Olah et~al.(2018)Olah, Satyanarayan, Johnson, Carter, Schubert, Ye,
  and Mordvintsev]{Olah2018}
Chris Olah, Arvind Satyanarayan, Ian Johnson, Shan Carter, Ludwig Schubert,
  Katherine Ye, and Alexander Mordvintsev.
\newblock {The Building Blocks of Interpretability}.
\newblock \emph{Distill}, 3\penalty0 (3):\penalty0 e10, mar 2018.
\newblock URL \url{https://distill.pub/2018/building-blocks}.

\bibitem[Paszke et~al.(2017)Paszke, Gross, Chintala, Chanan, Yang, DeVito, Lin,
  Desmaison, Antiga, and Lerer]{Paszke2017}
Adam Paszke, Sam Gross, Soumith Chintala, Gregory Chanan, Edward Yang, Zachary
  DeVito, Zeming Lin, Alban Desmaison, Luca Antiga, and Adam Lerer.
\newblock Automatic differentiation in pytorch.
\newblock 2017.

\bibitem[Pedregosa et~al.(2011)Pedregosa, Varoquaux, Gramfort, Michel, Thirion,
  Grisel, Blondel, Prettenhofer, Weiss, Dubourg, Vanderplas, Passos,
  Cournapeau, Brucher, Perrot, and Duchesnay]{Pedregosa2011}
Fabian Pedregosa, Ga{\"{e}}l Varoquaux, Alexandre Gramfort, Vincent Michel,
  Bertrand Thirion, Olivier Grisel, Mathieu Blondel, Peter Prettenhofer, Ron
  Weiss, Vincent Dubourg, Jake Vanderplas, Alexandre Passos, David Cournapeau,
  Matthieu Brucher, Matthieu Perrot, and {\'{E}}douard Duchesnay.
\newblock {Scikit-learn: Machine Learning in Python}.
\newblock \emph{Journal of Machine Learning Research}, 12\penalty0
  (Oct):\penalty0 2825--2830, 2011.
\newblock URL \url{http://www.jmlr.org/papers/v12/pedregosa11a}.

\bibitem[Ribeiro et~al.(2016)Ribeiro, Singh, and Guestrin]{Ribeiro2016}
Marco~Tulio Ribeiro, Sameer Singh, and Carlos Guestrin.
\newblock {"Why Should I Trust You?"}.
\newblock In \emph{Proceedings of the 22nd ACM SIGKDD International Conference
  on Knowledge Discovery and Data Mining - KDD '16}, pp.\  1135--1144, New
  York, New York, USA, 2016. ACM Press.
\newblock URL \url{http://dl.acm.org/citation.cfm?doid=2939672.2939778}.

\bibitem[Roeser \& Baier(2003)Roeser and Baier]{Roeser2003}
Tobias Roeser and Herwig Baier.
\newblock {Visuomotor Behaviors in Larval Zebrafish after GFP-Guided Laser
  Ablation of the Optic Tectum}.
\newblock 2003.
\newblock URL \url{http://www.jneurosci.org/content/jneuro/23/9/3726.full.pdf}.

\bibitem[Semmelhack et~al.(2014)Semmelhack, Donovan, Thiele, Kuehn, Laurell,
  and Baier]{Semmelhack2014}
Julia~L Semmelhack, Joseph~C Donovan, Tod~R Thiele, Enrico Kuehn, Eva Laurell,
  and Herwig Baier.
\newblock {A dedicated visual pathway for prey detection in larval zebrafish}.
\newblock 3:\penalty0 4878, 2014.
\newblock URL
  \url{https://cdn.elifesciences.org/articles/04878/elife-04878-v3.pdf}.

\bibitem[Shin et~al.(2016)Shin, Roth, Gao, Lu, Xu, Nogues, Yao, Mollura, and
  Summers]{Shin2016}
Hoo~Chang Shin, Holger~R. Roth, Mingchen Gao, Le~Lu, Ziyue Xu, Isabella Nogues,
  Jianhua Yao, Daniel Mollura, and Ronald~M. Summers.
\newblock {Deep Convolutional Neural Networks for Computer-Aided Detection: CNN
  Architectures, Dataset Characteristics and Transfer Learning}.
\newblock \emph{IEEE Transactions on Medical Imaging}, 35\penalty0
  (5):\penalty0 1285--1298, may 2016.
\newblock URL \url{http://ieeexplore.ieee.org/document/7404017/}.

\bibitem[Simonyan \& Zisserman(2014)Simonyan and Zisserman]{Simonyan2014}
Karen Simonyan and Andrew Zisserman.
\newblock {Two-Stream Convolutional Networks for Action Recognition in Videos},
  2014.
\newblock URL \url{http://papers.nips.cc/paper/5353-two-stream-convolutional}.

\bibitem[Simonyan et~al.(2014)Simonyan, Vedaldi, and Zisserman]{Simonyan2014b}
Karen Simonyan, Andrea Vedaldi, and Andrew Zisserman.
\newblock {Deep Inside Convolutional Networks: Visualising Image Classification
  Models and Saliency Maps}.
\newblock 2014.
\newblock URL \url{http://arxiv.org/abs/1312.6034}.

\bibitem[Springenberg et~al.(2015)Springenberg, Dosovitskiy, Brox, and
  Riedmiller]{Springenberg2015}
Jost~Tobias Springenberg, Alexey Dosovitskiy, Thomas Brox, and Martin
  Riedmiller.
\newblock {Striving for Simplicity: The All Convolutional Net}.
\newblock 2015.
\newblock URL \url{http://arxiv.org/abs/1412.6806}.

\bibitem[Temizer et~al.(2015)Temizer, Donovan, Baier, and
  Semmelhack]{Temizer2015}
Incinur Temizer, Joseph~C. Donovan, Herwig Baier, and Julia~L. Semmelhack.
\newblock {A Visual Pathway for Looming-Evoked Escape in Larval Zebrafish}.
\newblock \emph{Current Biology}, 25\penalty0 (14):\penalty0 1823--1834, jul
  2015.
\newblock URL
  \url{https://www.sciencedirect.com/science/article/pii/S0960982215006673}.

\bibitem[van~der Walt et~al.(2011)van~der Walt, Colbert, and
  Varoquaux]{VanDerWalt2011}
St{\'{e}}fan van~der Walt, S~Chris Colbert, and Ga{\"{e}}l Varoquaux.
\newblock {The NumPy Array: A Structure for Efficient Numerical Computation}.
\newblock \emph{Computing in Science {\&} Engineering}, 13\penalty0
  (2):\penalty0 22--30, mar 2011.
\newblock URL \url{http://ieeexplore.ieee.org/document/5725236/}.

\bibitem[{Van Horn} \& Perona(2017){Van Horn} and Perona]{VanHorn2017}
Grant {Van Horn} and Pietro Perona.
\newblock {The Devil is in the Tails: Fine-grained Classification in the Wild}.
\newblock 2017.
\newblock URL \url{www.inaturalist.org http://arxiv.org/abs/1709.01450}.

\bibitem[{Van Molle} et~al.(2018){Van Molle}, {De Strooper}, Verbelen,
  Vankeirsbilck, Simoens, and Dhoedt]{VanMolle2018}
Pieter {Van Molle}, Miguel {De Strooper}, Tim Verbelen, Bert Vankeirsbilck,
  Pieter Simoens, and Bart Dhoedt.
\newblock {Visualizing convolutional neural networks to improve decision
  support for skin lesion classification}.
\newblock In \emph{Lecture Notes in Computer Science (including subseries
  Lecture Notes in Artificial Intelligence and Lecture Notes in
  Bioinformatics)}, volume 11038 LNCS, pp.\  115--123, 2018.
\newblock URL \url{https://doi.org/10.1007/978-3-030-02628-8_13}.

\bibitem[White et~al.(2013)White, Rose, and Zon]{White2013}
Richard White, Kristin Rose, and Leonard Zon.
\newblock {Zebrafish cancer: The state of the art and the path forward}, 2013.
\newblock URL \url{www.nature.com/reviews/cancer}.

\bibitem[Zach et~al.(2007)Zach, Pock, and Bischof]{Zach2007}
C.~Zach, T.~Pock, and H.~Bischof.
\newblock {A Duality Based Approach for Realtime TV-L 1 Optical Flow}.
\newblock In \emph{Pattern Recognition}, pp.\  214--223. Springer Berlin
  Heidelberg, Berlin, Heidelberg, 2007.
\newblock URL \url{http://link.springer.com/10.1007/978-3-540-74936-3_22}.

\bibitem[Zeiler \& Fergus(2014)Zeiler and Fergus]{Zeiler2014}
Matthew~D Zeiler and Rob Fergus.
\newblock {Visualizing and Understanding Convolutional Networks}.
\newblock 2014.
\newblock URL \url{https://cs.nyu.edu/fergus/papers/zeilerECCV2014.pdf}.

\end{thebibliography}
\bibliographystyle{iclr2020_conference}

\appendix
\renewcommand{\thefigure}{S\arabic{figure}}
\section{Fitting the Baseline SVM} \label{app:svm}
\begin{figure}[ht]
	\begin{center}
		\includegraphics[width=0.35\textwidth]{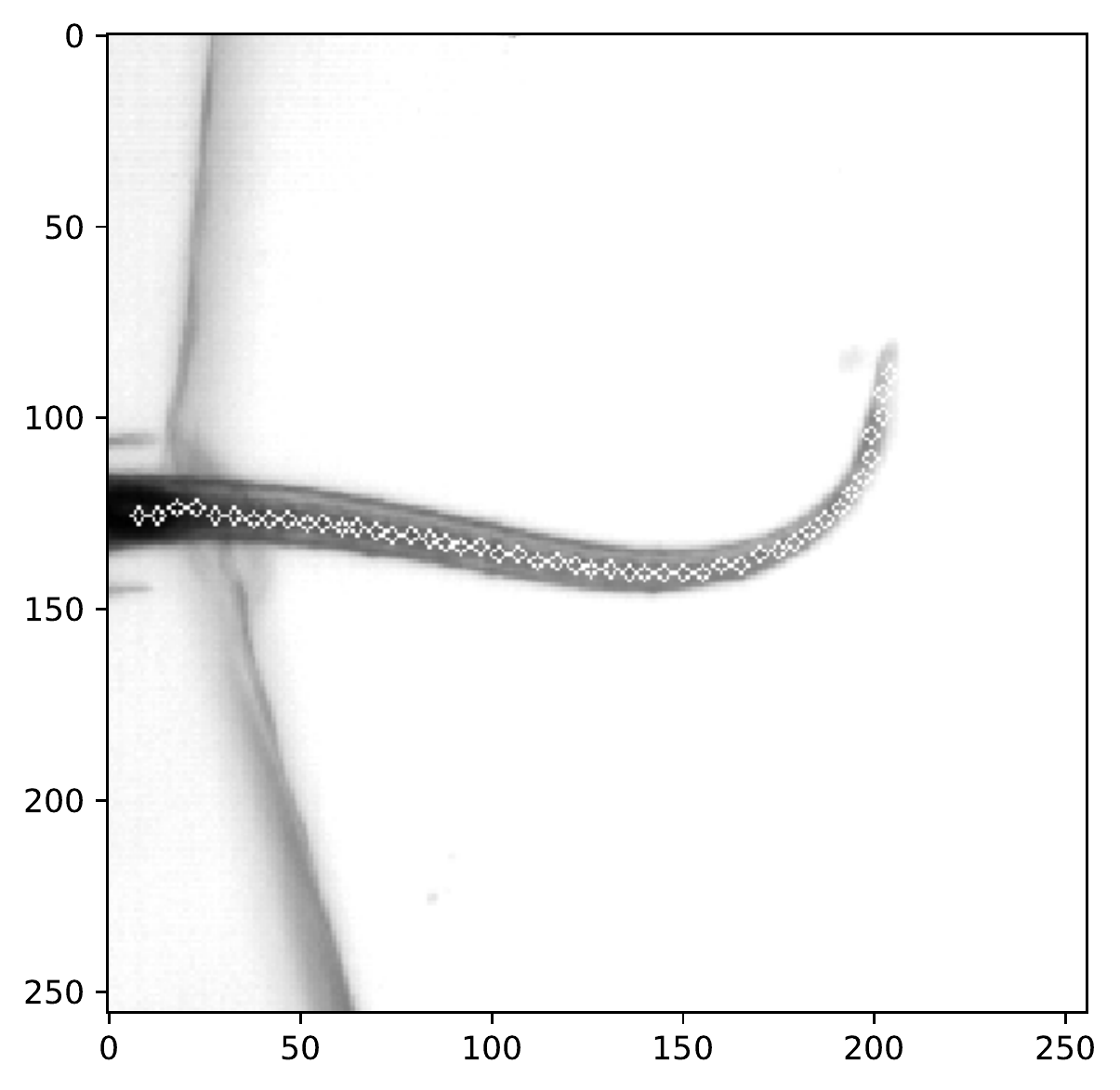}
		\caption{Distribution of points along the tail after running the tail-fitting procedure.}
		\label{fig:points}
	\end{center}
\end{figure}

% This section should go into the appendix.  In its place, you should include more CNN details. - Arno
For each frame in the 1,214 videos, we applied the tail-fitting code developed by \cite{Semmelhack2014} to compute points along the tail, as depicted in Figure~\ref{fig:points}.
We initialized their procedure central and 8 pixels from the left edge, because after pre-processing we could assume this to be just next to the right end of the bladder.
Some of the videos contained frames which the tail-fitting code had problems processing, possibly because the pre-processing procedure cut off the tip of the tail in these instances.
This resulted in 953 correctly processed videos, including 482 (50.6\%) spontaneous and 471 (49.4\%) prey bouts.
We performed no augmentation here because this would not have benefited the SVM.

The feature extraction and model fitting algorithm then split the set into 85\% training and 15\% held-out test sets.
\cite{Semmelhack2014} have identified 5 key features which allowed their SVM to achieve a cross-validation accuracy of 96\%.
They did not report results on a held-out test set.
We used their provided code to extract these features.
Then we performed a grid-search to tune SVM-kernel, $\gamma$, and C\footnote{RBF-kernel, $\gamma \in$ \{1e-1, 1e-2, 1e-3, 1e-4\}, C $\in$ \{0.01, 0.1, 1, 10\}; linear-kernel, C $\in$ \{0.01, 0.1, 1, 10\}.}.
Just like \cite{Semmelhack2014} we used stratified 5-fold cross-validation (\cite{Kohavi1995}).

\section{Project Pipeline} \label{app:details}
\begin{figure}[ht]
	\begin{center}
		\includegraphics[width=0.85\columnwidth]{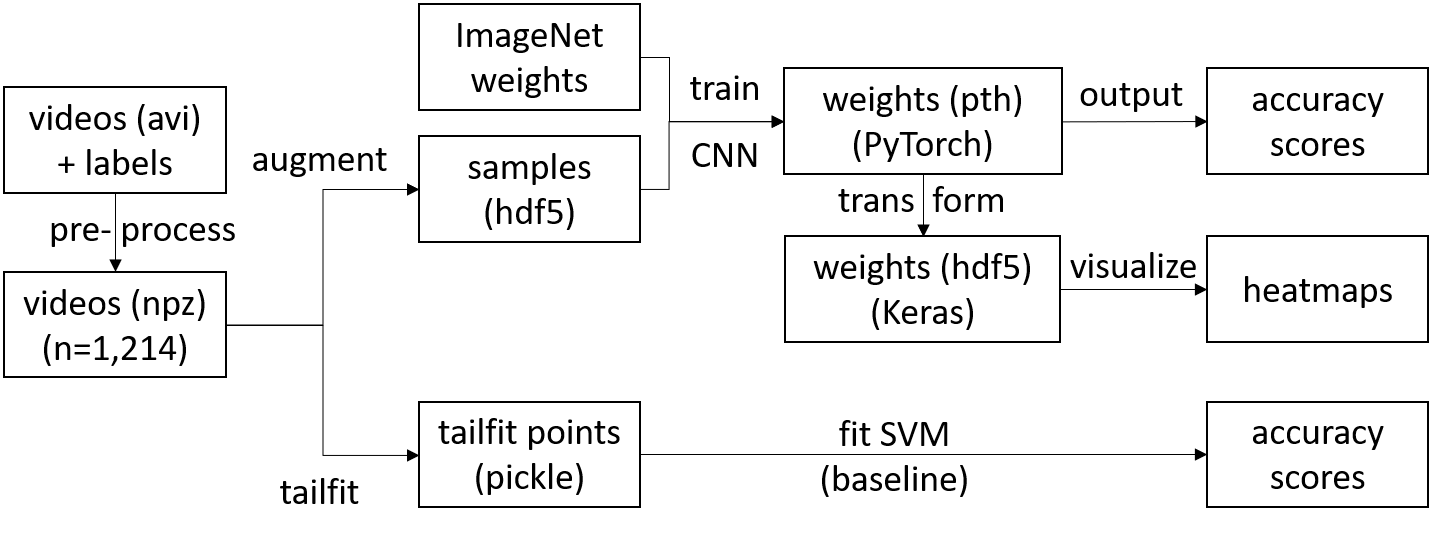}
		\caption{Overview over the whole project pipeline.}
		\label{fig:pipeline}
	\end{center}
\end{figure}

\begin{figure}[ht]
	\begin{center}
		\includegraphics[width=0.9\columnwidth]{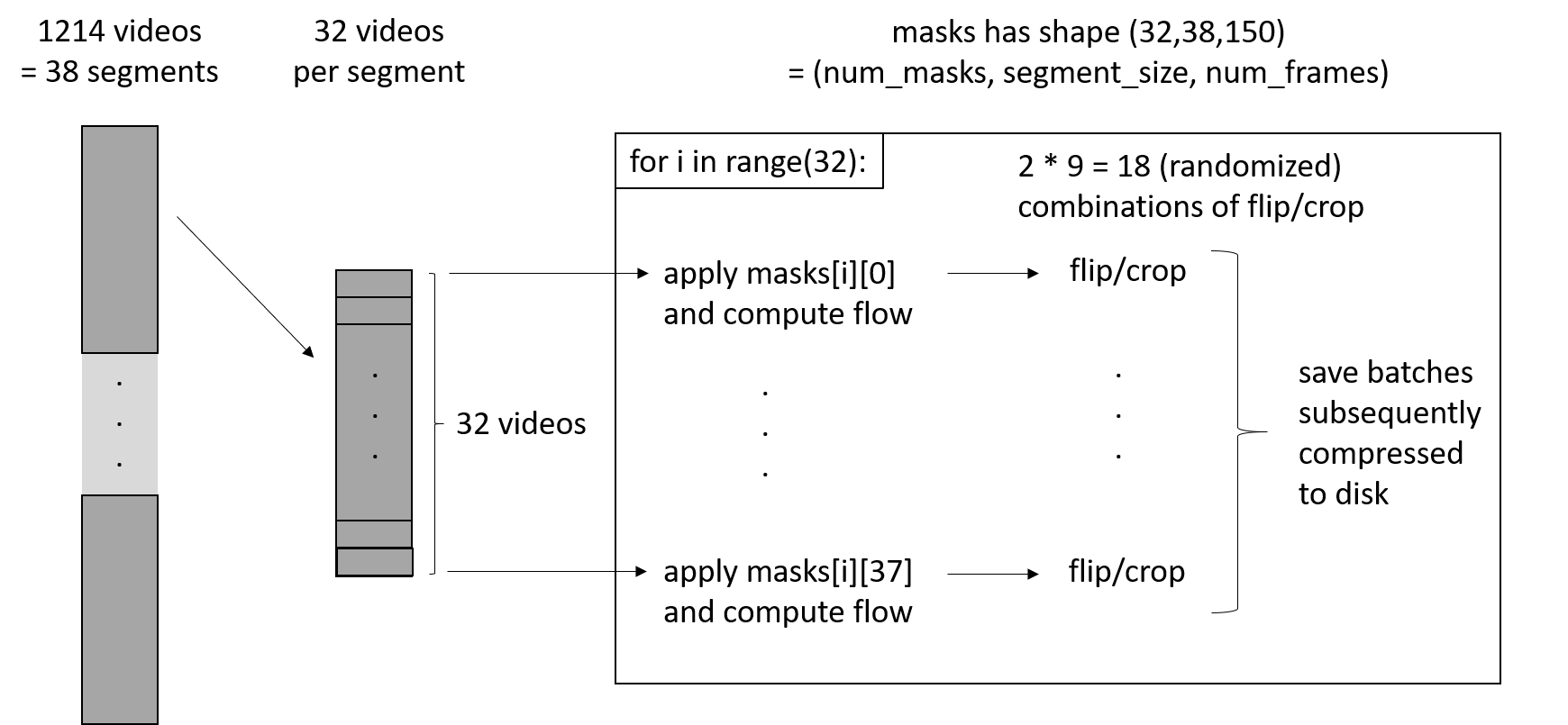}
		\caption{Data augmentation procedure.}
		\label{fig:aug}
	\end{center}
\end{figure}

Figure~\ref{fig:pipeline} gives an overview over the whole project pipeline. The figure includes a depiction of the raw data input, pre-processing, augmentation, CNN and SVM training, and heatmap analysis.
Figure~\ref{fig:aug} summarizes the data augmentation procedure.
All scripts worked with a seed of $462019$.
% Ideally, version numbers should be added. - Arno
We used the openly available distributions of NumPy~1.16.4 (\cite{VanDerWalt2011}), Matplotlib~3.1.1 (\cite{Hunter2007}), tqdm~4.32.2 (\cite{DaCosta-Luis2019}), OpenCV~4.1.0.25 (\cite{opencv_library}),  scikit-learn~0.21.2 (\cite{Pedregosa2011}), PyTorch~1.1.0 (\cite{Paszke2017}), h5py~2.9.0 (\cite{hdf5}), TensorFlow~1.14.0 (\cite{tensorflow2015}), Keras~2.2.4 (\cite{keras}), and iNNvestigate~1.0.8 (\cite{Alber2018}).

\textbf{Pre-processing.}
% You should start with a sentence summarizing the goal of this pre-processing. - Arno
We aimed to center the fish's bladder on the left of each cropped frame.
To achieve this, we applied a binary threshold at value $3$, because after normalization and gamma correction the pixels of the bladder were separated mostly below this value.
While the fish was quite light, the eyes as the second darkest part of the fish might still have fallen under this threshold.
Therefore, we first detected all contours to discard tiny contours ($< 0.01\%$ of the whole frame) and then kept only the right-most contour.
Since this had to be the bladder now, we could get the crop dimensions using the right-most pixel of that contour.

Each raw video mainly consisted of a still fish interspersed with a few short bouts.
These were the events we extracted into consecutive 150 frames each.
The idea was to detect motion by checking the percentage of pixel value changes from one frame to the next, considering only the tail.
We omitted pixels other than the tail with a simple binary threshold at value 200.
The pixels had to change in our case at least 0.38\% of the entire pixel range ($\text{height} \times \text{width} \times 255$) in order for motion to be detected.
If the algorithm detected motion for a certain number of consecutive frames, it set this as the start of an event.
Also, it added a preceding buffer of 15 frames.
The end was set 150 frames from the start.
If no motion was detected, we automatically took the first frame as a start.

We had to take care that extracted videos did not overlap, i.e. in part contained identical frames, even if there were two or more distinct movements contained within 150 frames.
This might have otherwise lead to train/test-contamination.
Therefore, we discarded any detected motions which fell in the range of a previous video.
One special case we did not take into account was when the start was less than 150 frames from the end of the file.
We shifted the start back by just enough frames to fit it in, but this might have made it overlap with a previous video.
Since this case was probably very rare and not detrimental, we have kept the code as it was.

\textbf{Data augmentation.}
% You can remove the next sentence or move it to the appendix. - Arno
We parallelized data augmentation with 38 workers, because optical flow calculation took roughly 14 seconds per subsample.
% You can remove the next sentence or move it to the appendix. - Arno
We used a Dell PowerEdge R815 with four 16 core Opteron CPUs, 256~GB memory, and 4~TB of disk space (HDD).
% The rest of this paragraph can go into appendix. - Arno
Furthermore, the resulting hdf5-files were compressed with gzip-compression at maximum compression level.
This outperformed lzf-copmression by a factor of 1.76 when comparing training times.
This advantage can be ascribed to heavy parallelization of decompression and relatively low transfer speeds between hard drive and memory.
% The rest of this paragraph should go into appendix. - Arno

\textbf{Training procedure.}
We implemented PyTorch's \code{Dataset} module to make use of the multiprocessing capabilities of the \code{DataLoader} class on servers with 16--32 CPU cores and 4 GPUs (NVIDIA GTX1060 6~GB) each.
This was necessary to achieve manageable epoch times.

\textbf{Relevance analysis with heatmaps.}
The toolbox ``iNNvestigate'' (\cite{Montavon2017}) for analyzing the trained weights only supported Keras with the TensorFlow-backend.
Therefore, we re-implemented the exact structure of our CNN and initialized it with the extracted weights from PyTorch.
While the conversion could have been done with tools like ONNX, after a few unsuccessful attempts we transported the weights with a custom Python script.
% The beginning of this paragraph up to this point should go into appendix. - Arno

% Do you have an intuition for why the toolbox had problems with some of the samples?  You should definitely elaborate here.  If this is technical then this whole paragraph should go into appendix. - Arno
% Unfortunately, I could not find any rule for which samples were unsuccessfully analyzed. - Bennet
A caveat to the ``iNNvestigate'' toolbox emerged after heatmap generation: it had problems analyzing 1,578 of the 3,420 samples.
These produced an empty output.
We made sure the problematic samples did not follow any systematic pattern by checking the indices of correctly analyzed samples, the ratio of true positives and negatives and false positives and negatives, as well as the distribution of classification confidence after the analysis.
Since all numbers were the same as before the analysis, we continued with 1,842 samples for further investigation.

\section{Other Explainability Techniques} \label{app:sens}
\begin{figure}[ht]
	\centering
	\begin{subfigure}{.55\textwidth}
		\centering
		\includegraphics[width=\linewidth]{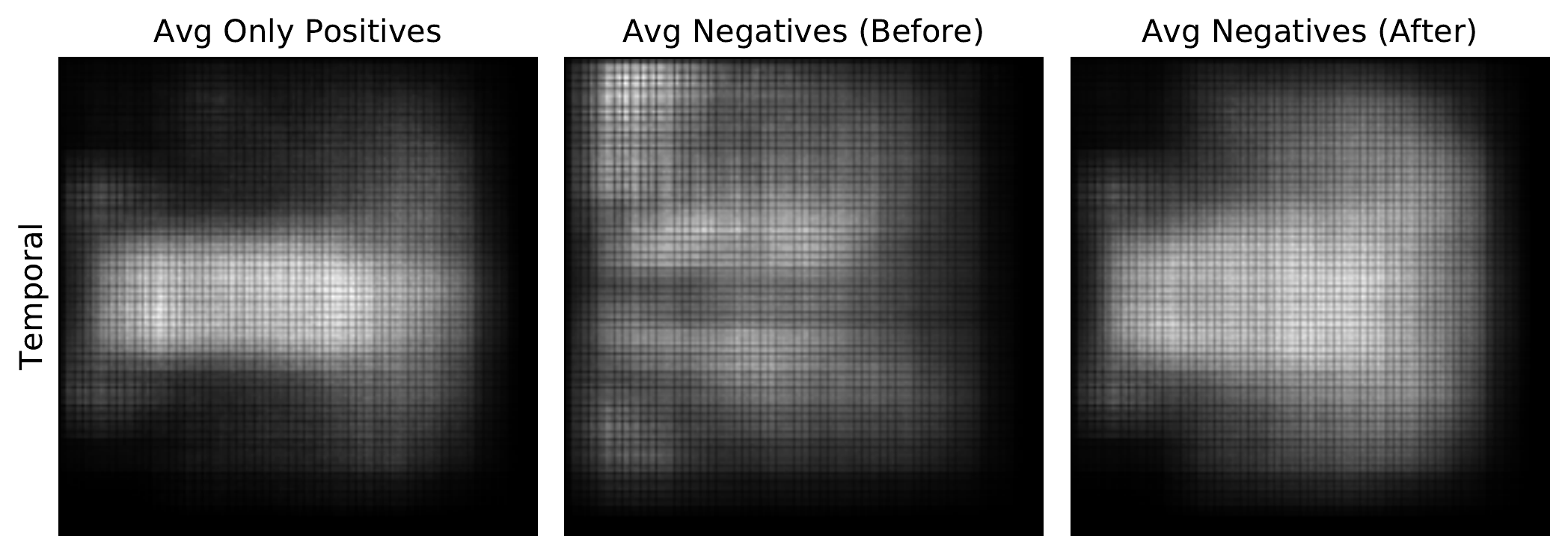}
	\end{subfigure}
	\begin{subfigure}{.22\textwidth}
		\vskip 5pt
		\centering
		\includegraphics[width=\linewidth]{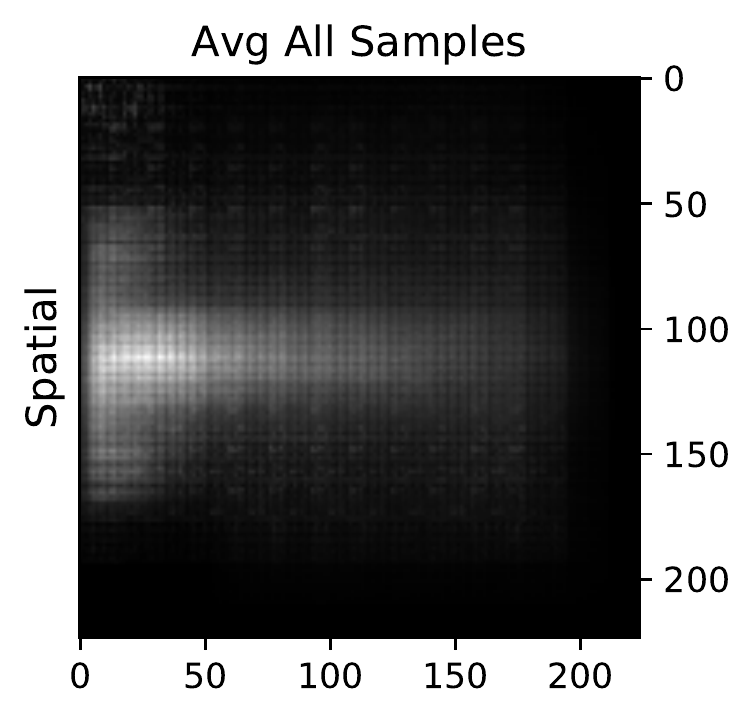}
	\end{subfigure}
	\caption{
		Averaged saliency maps, analogous to Figure~\ref{fig:the_average_frame}.}
	\label{fig:sens}
\end{figure}

\begin{figure}[ht]
	\centering
	\begin{subfigure}{.55\textwidth}
		\centering
		\includegraphics[width=\linewidth]{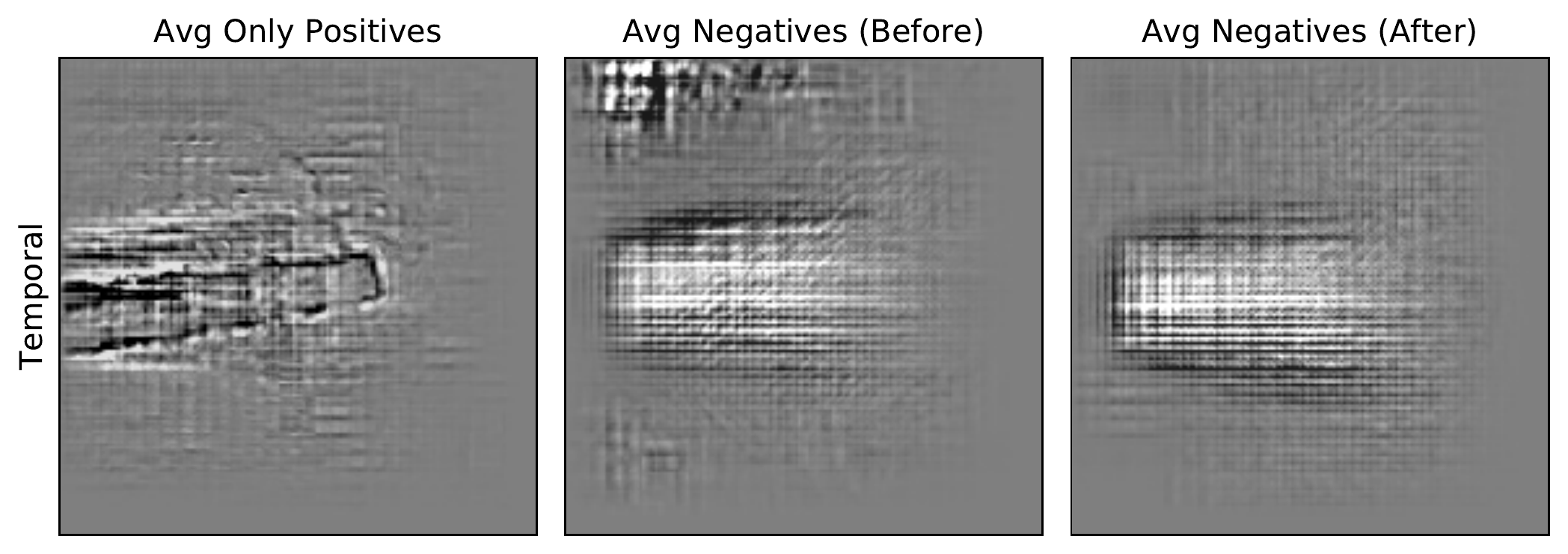}
	\end{subfigure}
	\begin{subfigure}{.22\textwidth}
		\vskip 5pt
		\centering
		\includegraphics[width=\linewidth]{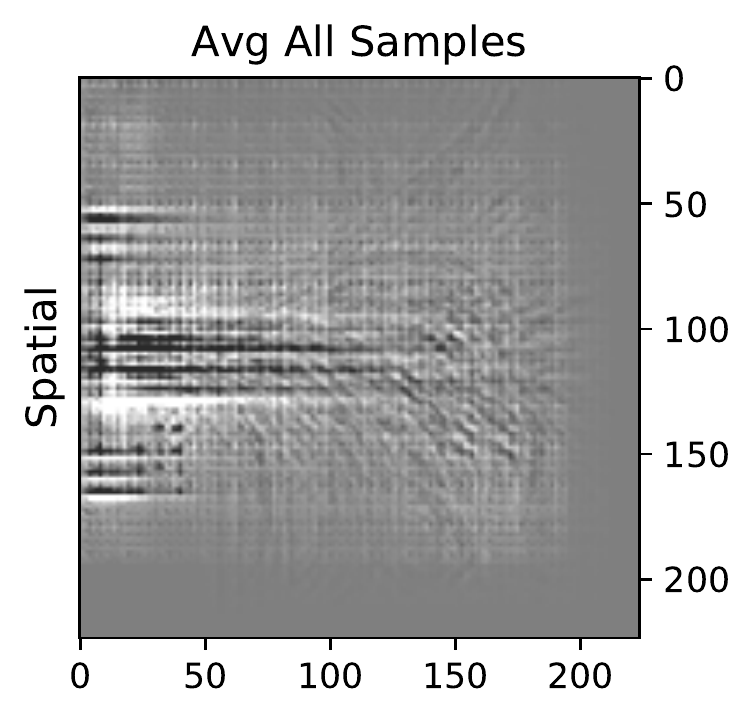}
	\end{subfigure}
	\caption{
		Averaged heatmaps from Guided BackProp, analogous to Figure~\ref{fig:the_average_frame}.}
	\label{fig:gui}
\end{figure}

We generated saliency maps (\cite{Simonyan2014b}) and Guided BackProp heatmaps (\cite{Springenberg2015}) analogous to the relevance heatmaps in Figures~\ref{fig:sens} and~\ref{fig:gui} for comparison with the more recent technique of DTD.
It becomes apparent that these other two techniques allow similar insights, although slightly fuzzier.
Importantly, they also uncover the ``Clever Hans'' prediction.

\section{Selection of Additional Heatmaps}
\begin{figure}[ht]
	\begin{center}
		\includegraphics[width=\columnwidth]{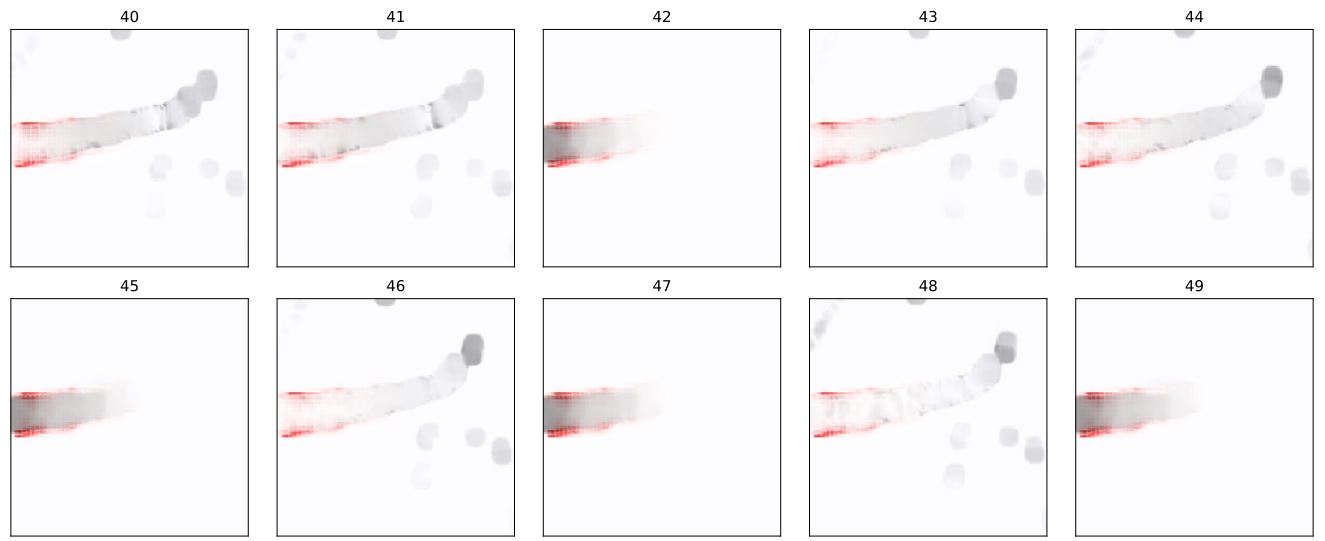}
		\caption{Selection of flow frames of the most confident true positive sample.}
		\label{fig:tp}
	\end{center}
\end{figure}

\begin{figure}[ht]
	\begin{center}
		\includegraphics[width=\columnwidth]{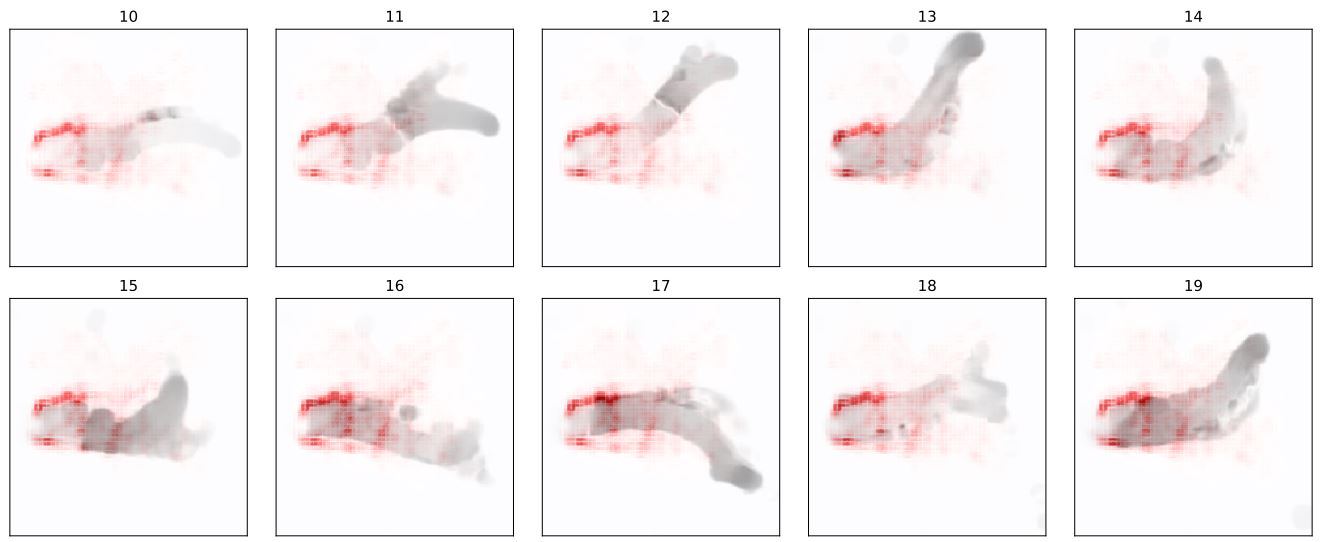}
		\caption{Selection of flow frames of the most confident true negative sample.}
		\label{fig:tn}
	\end{center}
\end{figure}

\begin{figure}[ht]
	\begin{center}
		\includegraphics[width=\columnwidth]{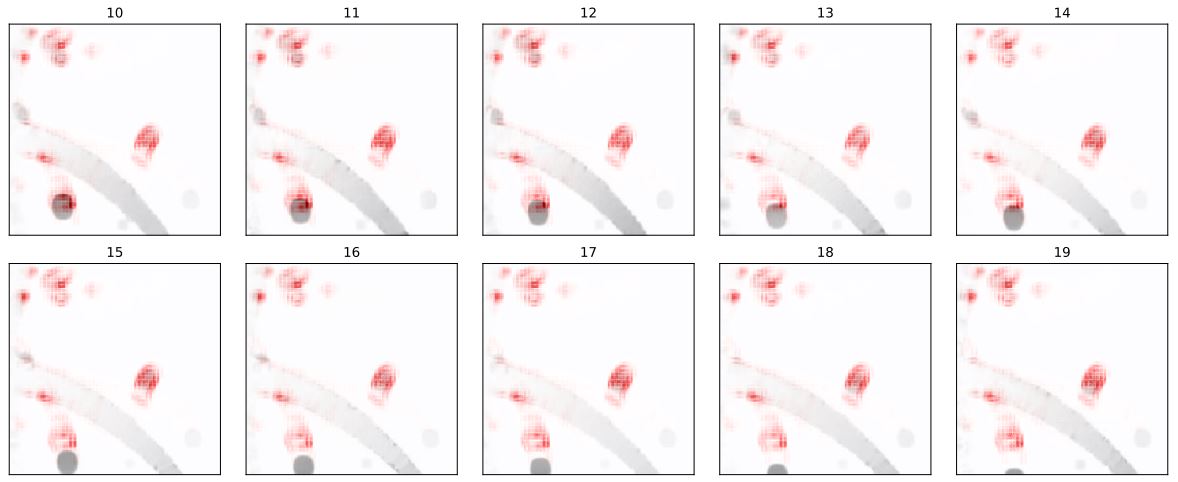}
		\caption{Selection of flow frames of the most confident false positive sample.}
		\label{fig:fp}
	\end{center}
\end{figure}

\begin{figure}[ht]
	\begin{center}
		\includegraphics[width=\columnwidth]{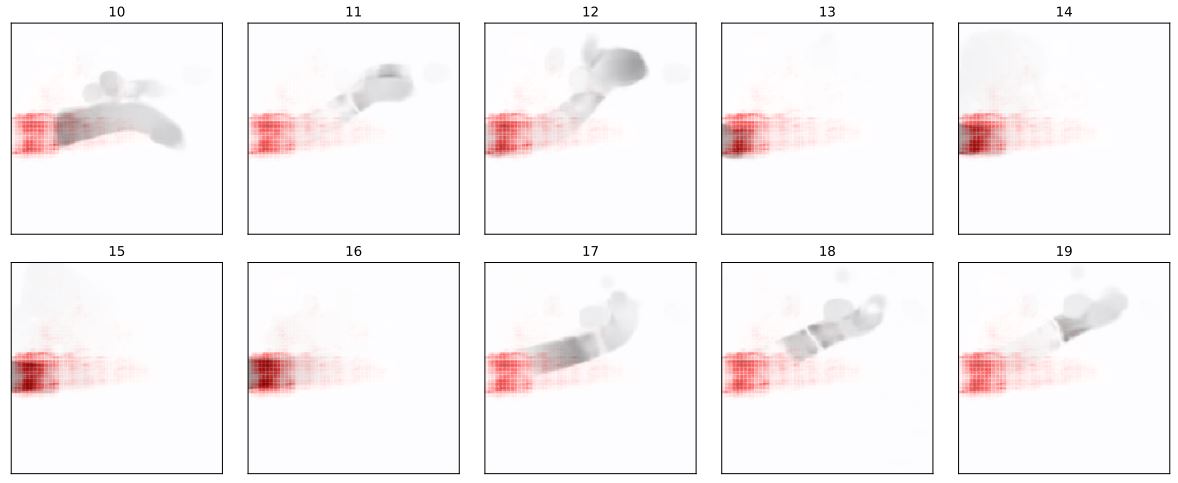}
		\caption{Selection of flow frames of the most confident false negative sample.}
		\label{fig:fn}
	\end{center}
\end{figure}

\begin{figure}[ht]
	\begin{center}
		\includegraphics[width=\columnwidth]{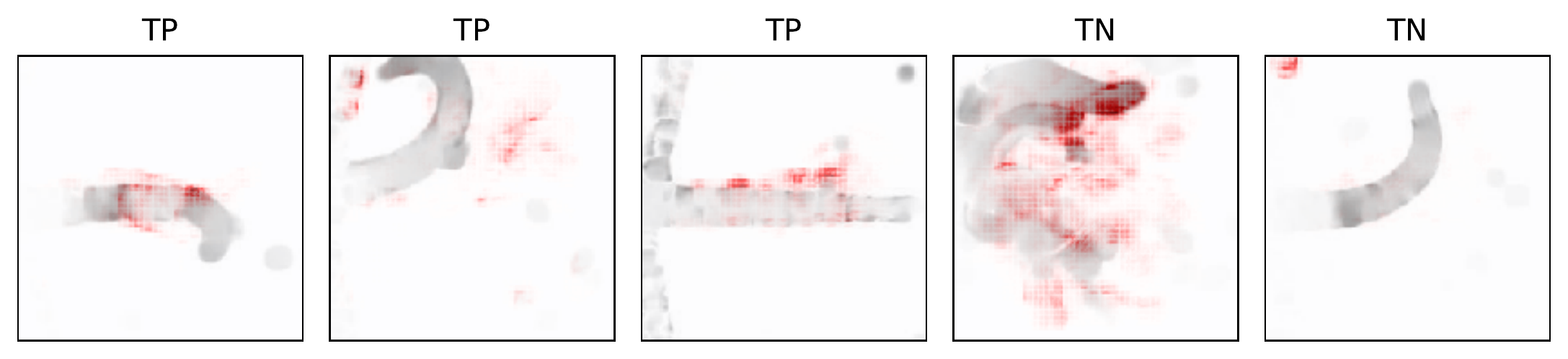}
		\caption{Selection of flow frames of a selection of true positive (TP) and true negative samples (TN).}
		\label{fig:sel_temporal}
	\end{center}
\end{figure}

\begin{figure}[ht]
	\begin{center}
		\includegraphics[width=0.9\columnwidth]{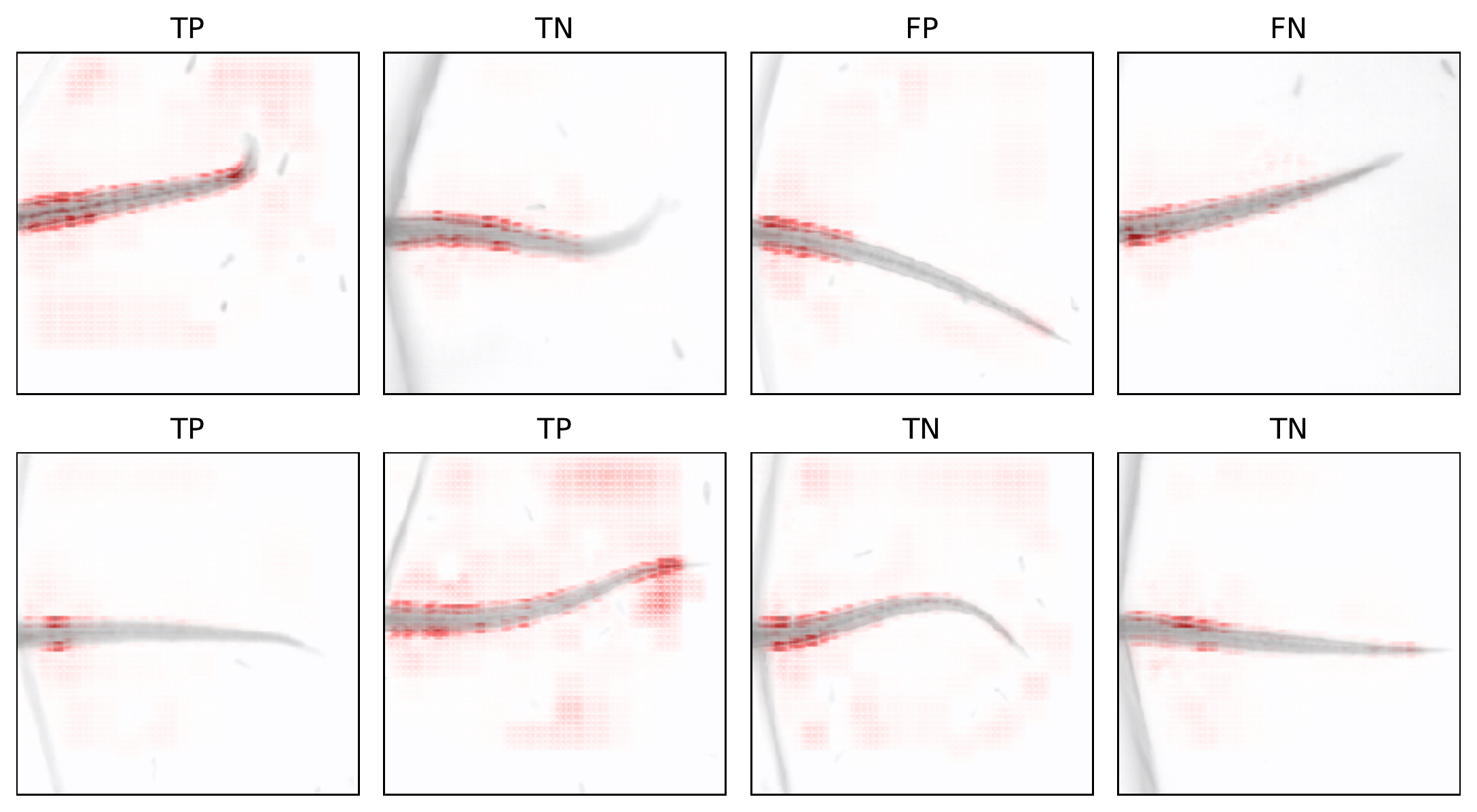}
		\caption{Upper row: spatial input of the most confident samples of each category. Lower row: spatial input of a selection of samples.}
		\label{fig:sel_spatial}
	\end{center}
\end{figure}

Here, we depict the ten most informative consecutive flow frames of the single most confident true positive (Figure~\ref{fig:tp}), true negative (Figure~\ref{fig:tn}), false positive (Figure~\ref{fig:fp}), and false negative (Figure~\ref{fig:fn}) sample.
Figure~\ref{fig:sel_spatial} summarizes the spatial heatmaps of the same samples.
Moreover, we gather five particularly informative flow frames in Figure~\ref{fig:sel_temporal} and four spatial heatmaps in Figure~\ref{fig:sel_spatial}.


\section{Confidence Analysis on Heatmaps}
\begin{figure}[ht]
	\begin{center}
		\includegraphics[width=\columnwidth]{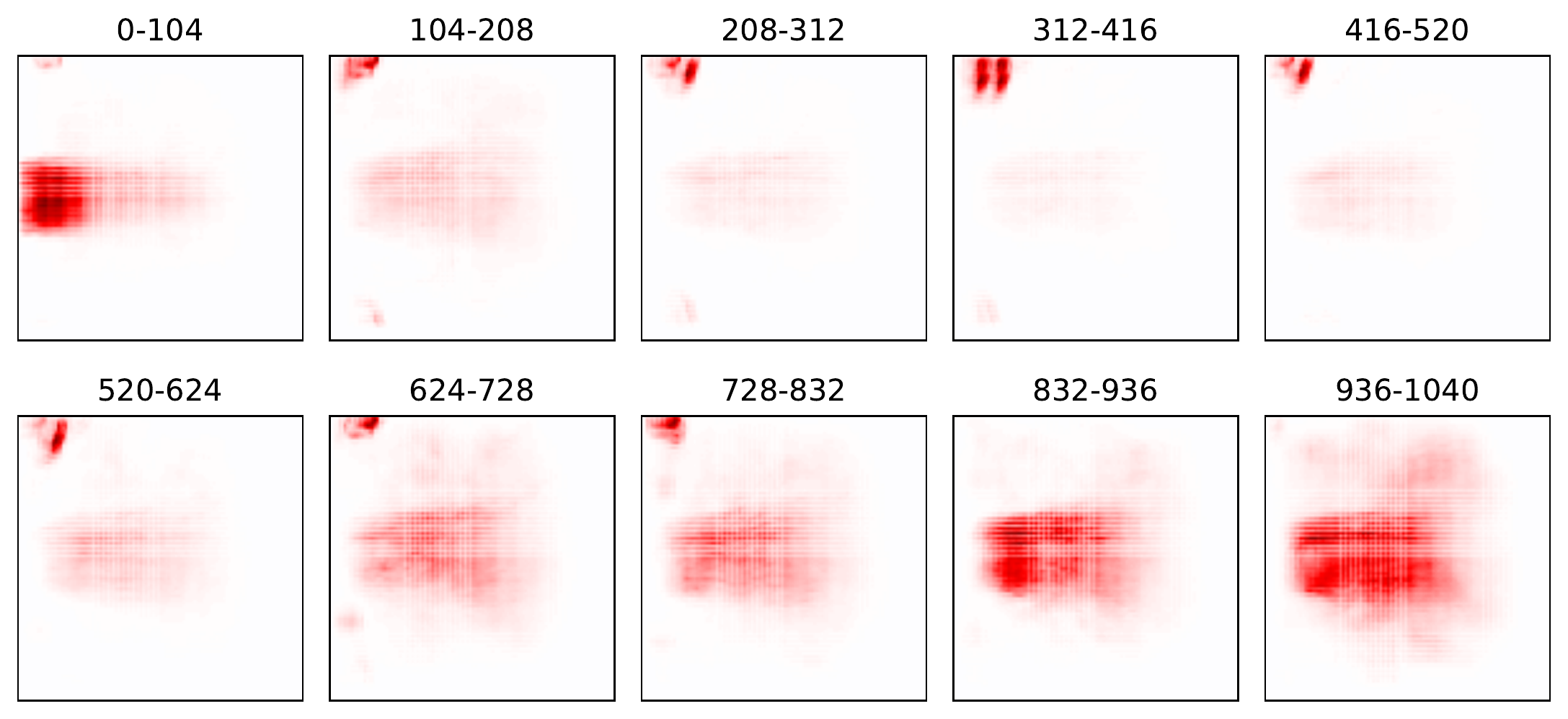}
		\caption{Averaged heatmaps of 104 samples per subfigure, sorted by increasing confidence for responding negative.}
		\label{fig:agarose}
	\end{center}
\end{figure}

We performed a confidence analysis on the heatmaps which depicted the ``Clever Hans'' feature of agarose motion in the top left corner.
We sorted all negative classifications by increasing confidence calculated as 
$-\log(\nicefrac{\log(\text{P}(1))}{\log(\text{P}(0))})$ for each sample.
We then grouped and averaged the heatmaps over windows of 104 samples, as shown in Figure~\ref{fig:agarose}.
The analysis uncovered that the more confident a negative classification, the more the CNN relied on tail features.
This in turn indicated that the CNN was able to learn actual features on the tail and did not entirely rely on agarose motion.
Also, it suggested that tail features were a better predictor than agarose motion.

\end{document}